\documentclass[]{single_column_tech_report}

\usepackage[utf8]{inputenc}
\usepackage{xspace}
\usepackage{url}

\usepackage{amsmath}
\usepackage{amssymb}
\usepackage{amsfonts}
\usepackage{amsthm}
\usepackage{mathtools}
\usepackage{mathrsfs}

\usepackage{adjustbox}
\usepackage{wrapfig}

% ...
\usepackage{array}
\usepackage{multicol}
\usepackage{makecell}
\usepackage{tabularx}
\newcolumntype{Y}{>{\raggedright\arraybackslash}X}
\newcolumntype{Z}{>{\raggedleft\arraybackslash}X}
\newcolumntype{L}[1]{>{\raggedright\arraybackslash}p{#1}}
\newcolumntype{C}[1]{>{\centering\arraybackslash}p{#1}}
\newcolumntype{R}[1]{>{\raggedleft\arraybackslash}p{#1}}

\usepackage{algorithm}
\usepackage{algpseudocode}

\usepackage{changepage}
\usepackage{anyfontsize}
\usepackage{marvosym}
\usepackage[normalem]{ulem}
\usepackage{listings}

\lstdefinestyle{promptlisting}{
  basicstyle=\ttfamily\scriptsize,
  breaklines=true,
  breakatwhitespace=false,
  columns=fullflexible,
  keepspaces=true,
  showstringspaces=false,
  frame=single,
  rulecolor=\color{blockborder},
  backgroundcolor=\color{blockbg},
  xleftmargin=4pt,
  xrightmargin=4pt,
  aboveskip=6pt,
  belowskip=7pt
}

\renewcommand{\titlefont}{%
  \fontsize{20}{24}\selectfont
}

\renewcommand{\title}[1]{%
  \gdef\titlelist{%
    {\titlefont\sffamily\bfseries #1}%
  }%
}

\definecolor{lightgreen}{RGB}{144,238,144}
\definecolor{deepgreen}{RGB}{0,136,55}

\definecolor{blockbg}{HTML}{F0F4F8}
\definecolor{blockborder}{HTML}{B0C4D8}
\definecolor{actioncolor}{HTML}{2B6CB0}
\definecolor{obscolor}{HTML}{4A5568}
\definecolor{commentcolor}{HTML}{718096}

\definecolor{tblInk}{HTML}{000000}
\definecolor{tblRule}{HTML}{000000}
\definecolor{tblHead}{HTML}{FFFFFF}
\definecolor{tblGroup}{HTML}{FFFFFF}
\definecolor{tblOurs}{HTML}{E6E6FF}
\definecolor{tblStripe}{HTML}{FAFAFA}
\definecolor{tblSummary}{HTML}{F2F2F2}
\definecolor{tblMuted}{HTML}{666666}
\definecolor{tblPositive}{HTML}{EAF4EE}
\definecolor{tblNegative}{HTML}{F8EEEE}

\definecolor{catVS}{HTML}{FCE4E4}
\definecolor{catMath}{HTML}{FFF4D6}
\definecolor{catGen}{HTML}{E0EFFF}
\definecolor{catAvg}{HTML}{EEEEEE}
\definecolor{rankTop}{HTML}{F8C8C8}
\definecolor{rankSnd}{HTML}{C8E8C8}
\definecolor{rankThd}{HTML}{C8D8F0}

\newenvironment{normaltable}{%
  \begingroup
  \small
  \setlength{\tabcolsep}{5pt}%
  \renewcommand{\arraystretch}{1.14}%
  \arrayrulecolor{tblRule}%
}{\endgroup}

\newenvironment{densetable}{%
  \begingroup
  \footnotesize
  \setlength{\tabcolsep}{3.5pt}%
  \renewcommand{\arraystretch}{1.10}%
  \arrayrulecolor{tblRule}%
}{\endgroup}

\newenvironment{prosetable}{%
  \begingroup
  \small
  \setlength{\tabcolsep}{5pt}%
  \renewcommand{\arraystretch}{1.18}%
  \arrayrulecolor{tblRule}%
}{\endgroup}

\newcommand{\tblhead}[1]{{\sffamily\bfseries\color{tblInk}#1}}
\newcommand{\tblbest}[1]{{\bfseries #1}}
\newcommand{\tblheaderrow}{\rowcolor{tblHead}}
\newcommand{\tblgrouprow}{\rowcolor{tblGroup}}
\newcommand{\tbloursrow}{\rowcolor{tblOurs}}

\newcommand{\tblnote}[1]{%
  \par\vspace{3pt}{\footnotesize\color{tblMuted}\raggedright #1\par}%
}

\newtcolorbox{trajectoryblock}[1][]{%
  enhanced,
  breakable,
  colback=blockbg,
  colframe=blockborder,
  boxrule=0.8pt,
  arc=3pt,
  left=6pt,
  right=6pt,
  top=5pt,
  bottom=5pt,
  fonttitle=\bfseries\small,
  title={#1},
  coltitle=black,
  attach boxed title to top left={yshift=-2mm, xshift=4mm},
  boxed title style={
    colback=blockbg,
    colframe=blockborder,
    boxrule=0.5pt,
    arc=2pt
  },
}

\hypersetup{
    colorlinks=true,
    linkcolor=deepgreen,
    citecolor=deepgreen,
    urlcolor=actioncolor
}

\newcommand{\bestval}[1]{\textbf{#1}}
\newlength{\acctokaccwidth}
\newlength{\acctoktokwidth}
\usepackage{fontawesome5}
\definecolor{spadecolor}{rgb}{1 , 0.7 , 0}
\definecolor{clubcolor}{rgb}{0.9 , 0.3 , 0.4}
\definecolor{heartcolor}{rgb}{0.8 , 0.4 , 0.9}
\definecolor{diamondcolor}{rgb}{0. , 0.5 , 0.9}
\definecolor{circlecolor}{rgb}{0.3 , 0.9 , 0.7}
\definecolor{squarecolor}{rgb}{0.5, 0.5, 0.5}

\definecolor{upgreen}{HTML}{32CD32}
\definecolor{downred}{HTML}{DC143C}

\definecolor{nicegreen}{RGB}{34,174,24}
\newcommand{\cmark}{{\color{nicegreen} \ding{51}}}
\newcommand{\xmark}{{\color{clubcolor} \ding{55}}}

\colorlet{tabtop1purple}{tblOurs}
\colorlet{tabtop1}{tblSummary}

\usepackage{pifont}
\usepackage{tabularx}
\usepackage{array}

\definecolor{nicegreen}{RGB}{34,174,24}
\definecolor{clubcolor}{rgb}{0.9,0.3,0.4}

\newcommand{\recovermark}{\xmark\,\ensuremath{\rightarrow}\,\cmark}

\title{VLM-in-Sandbox: Visual Workspaces for Agentic Visual Reasoning}

\author[1,2]{Hexiong Yang}
\author[1,3,4]{Mingrui Chen}
\author[1]{Jie Cao}
\author[1]{Ran He}

\affiliation[1]{NLPR\&MAIS, Institute of Automation, Chinese Academy of Sciences}
\affiliation[2]{School of Advanced Interdisciplinary Science, University of Chinese Academy of Sciences}
\affiliation[3]{School of Artificial Intelligence, University of Chinese Academy of Sciences}
\affiliation[4]{Zhongguancun Academy}

\abstract{
    Sandboxed computer environments support multi-step reasoning with tools, executable programs, and persistent files, yet their extension from language models to vision-language models (VLMs) introduces a distinct state-management problem. Visual reasoning produces intermediate image-valued evidence---crops, masks, overlays, zoomed regions, and analytic renderings---that must remain addressable without accumulating unboundedly in multimodal context.
    We introduce VLM-in-Sandbox, a training-free framework for agentic multimodal reasoning in controlled computer environments. Its Visual Workspace registers generated artifacts in an image ledger, maintains a bounded active visual context, and lets the model explicitly promote selected evidence for subsequent inspection. This separates visual evidence generation, performed by sandbox tools, from visual evidence management.
    Across seven benchmarks and four base VLMs, VLM-in-Sandbox achieves the highest sample-weighted average accuracy among Vanilla VLM, Append-only Sandbox, and the proposed method. A compiler-matched $2\times2$ study on 1,260 examples further separates model-directed visibility from bounded retention: VLM-in-Sandbox reaches 66.27\% accuracy with 18.6\% fewer total tokens than the automatic, retain-all control. Over all 6,350 submitted GPT-4.1-mini examples, it produces 302 rescues and 142 regressions relative to Original Append-only. A local vLLM study with prefix caching confirms that the smaller request workload also reduces uncached tokens, time to first token, and end-to-end latency. These results identify explicit visual evidence state as a central abstraction for sandboxed VLM agents.
}

\date{September 2026}

\begin{document}

\maketitle

\section{Introduction}

\label{sec:introduction}

LLM agents shift language models from passive predictors into interactive problem solvers that maintain state, invoke tools, and revise plans across multiple steps~\cite{react,webarena,osworld,visualwebarena}.
Sandboxed computer environments are a natural substrate for this behaviour, exposing a filesystem, shell commands, editing tools, persistent files, and executable programs rather than a single Python interpreter; recent LLM-in-Sandbox work shows they elicit general agentic behaviours that single-turn prompting cannot~\cite{llm_in_sandbox}.
A natural question follows: what additional abstractions are needed to extend this paradigm to vision-language models?

The extension is not a model substitution.
In language-centric agents, intermediate evidence is symbolic---code outputs, logs, file contents---all linearisable into the conversation.
A VLM agent often produces evidence that remains \emph{visual}: cropped regions, zoomed-in objects, rendered diagrams, analytic overlays, visual edits, or binary masks used to test hypotheses~\cite{visprog,vipergpt,visualsketchpad,refocus,pyvision,thyme,codev,openthinkimg,pyvisionrl}.
Generation alone does not make these usable: appended by default, they clutter the multimodal context; left as files, they remain invisible to the model.
VLMs do not need more tools; they need a mechanism to retain, index, select, and re-present visual evidence across the trajectory.
This need is especially acute as recent visual-reasoning systems increasingly scale inference, post-training, or action policies around intermediate visual operations~\cite{llamavo1,visualrft,reasonrft,visionr1,visuothink,vtoolr1,foveatedreasoning}.

\begin{figure}[t]
    \centering
    \includegraphics[width=\linewidth]{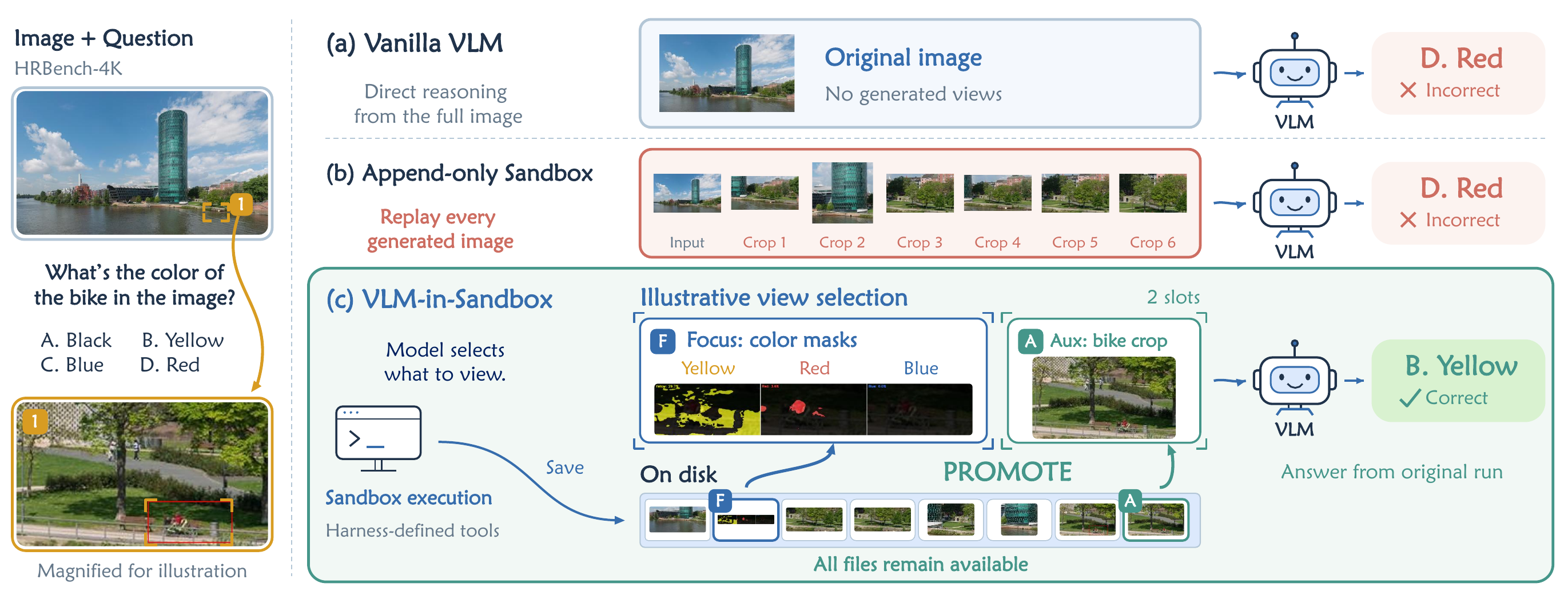}
    \caption{
    Motivating HRBench-4K example.
    \textbf{Vanilla VLM} reasons directly from the image; \textbf{Append-only Sandbox} accumulates generated views in its multimodal history; \textbf{VLM-in-Sandbox} keeps visual artifacts on disk and selects evidence for two active slots.
    The color-mask and bicycle-crop pairing illustrates view selection; the displayed answers are from the original runs, whose trace is discussed in \S\ref{sec:method}.
    }
    \label{fig:motivation}
\end{figure}

Figure~\ref{fig:motivation} illustrates the gap on a high-resolution visual-search instance from HRBench-4K~\cite{hrbench}.
A Vanilla VLM guesses from the downsampled image; Append-only Sandbox generates several crops but buries them in history; VLM-in-Sandbox promotes a curated subset into active visual context and answers correctly.
This motivates \emph{visual evidence state}: an addressable record of what evidence exists, where it came from, and which pieces should be exposed to the model at each step.

We propose \textsc{VLM-in-Sandbox}, a framework for agentic multimodal reasoning, and \textsc{Visual Workspace}, a runtime layer that organises the lifecycle of visual artifacts.
An \emph{Image Ledger} registers each generated image with a stable identifier, path, provenance, and parent--child links.
A bounded \emph{Active Visual Context}---two named slots, \texttt{focus} and \texttt{aux}---determines which artifacts are rendered as visual input next.
A \emph{Prompt Compiler} deterministically rebuilds each request from task, recent turns, ledger summary, and active context.
Crucially, generated artifacts do not enter the active context automatically: the model must issue an explicit \textsc{Promote} operation.
This enforces a clean separation between visual evidence \emph{generation} (sandbox) and \emph{management} (workspace).

We evaluate three inference modes---\textbf{Vanilla VLM}, \textbf{Append-only Sandbox}, and \textbf{VLM-in-Sandbox}---across seven multimodal benchmarks spanning three families:
high-resolution visual search (HRBench-4K~\cite{hrbench}, V$^{*}$Bench~\cite{vstar}),
multimodal mathematical reasoning (WeMath~\cite{wemath}, MathVision~\cite{mathvision}),
and general multimodal reasoning (MMStar~\cite{mmstar}, RealWorldQA~\cite{realworldqa}, MMMU~\cite{mmmu}).
Across four base VLMs, \textsc{VLM-in-Sandbox} achieves the best sample-weighted average accuracy among the three modes and reduces token use relative to Append-only Sandbox. Section~\ref{sec:experiments} presents the broad results, compiler-matched controls, and paired reliability analyses. Section~\ref{sec:workspace-analysis} examines context growth, prefix-cached inference cost, and qualitative examples. Together, these experiments test not only whether the full system works, but which visual-state choices account for its behavior and efficiency.

Our mechanism analysis shows that the workspace enables diverse computer-aided visual reasoning: models can render overlays, crop regions, compute masks, and use \textsc{Promote} to return selected evidence for later inspection. In summary, we identify visual evidence state as a missing abstraction for sandboxed VLM agents, instantiate it as the \textsc{Visual Workspace}, and evaluate it through broad, matched, paired, and systems-level evidence.

\section{Related Work}

\subsection{Sandboxed and Tool-Augmented Visual Reasoning}

\vspace{-0.3em} 
Recent agent systems place models in interactive web, desktop, and visually grounded environments rather than fixed single-turn traces~\cite{webarena,osworld,visualwebarena}. Vision-centric benchmarks further show that multimodal agents must coordinate perception, tool use, and long-horizon decisions~\cite{agentx,agenticmme,vtcbench}. Closest in spirit, LLM-in-Sandbox gives language models a general-purpose computer sandbox with filesystem access, terminal commands, editing tools, and outcome-based reinforcement learning~\cite{llm_in_sandbox}. Tool-augmented multimodal systems use visual modules, Python programs, sketches, image edits, expert calls, or learned tool policies to make visual reasoning more compositional~\cite{visprog,vipergpt,mmreact,chameleon,visualsketchpad,refocus,pyvision,thyme,codev,codedance,openthinkimg,pyvisionrl,deepeyes,vtoolr1}. These works establish the value of external computation, but most organize reasoning around a particular tool library, interpreter, canvas, or learned action policy. \textsc{VLM-in-Sandbox} instead asks how image-valued evidence produced inside a general computer workspace should persist, remain addressable, and return to the model.

\vspace{-0.3em} 

\subsection{Visual Evidence State and Evaluation}

\vspace{-0.3em} 

Active visual reasoning raises a state-management question: which visual evidence should remain visible after each step? Prior work studies guided search, visual working memory, small-detail localization, foveated focusing, compact visual abstractions, and latent visual memory~\cite{vstar,mllmsknow,visuothink,foveatedreasoning,sandboxvlm,vismem}. Visual Sketchpad maintains an episode-level visual canvas, while PyVision creates and revisits artifacts through a dynamic Python session~\cite{visualsketchpad,pyvision}; latent-memory methods retain information inside model state rather than as file-backed evidence~\cite{vismem}. VLM-in-Sandbox is complementary: it supplies a persistent filesystem, stable artifact identity and provenance, a bounded set of active visuals, and model-directed reactivation by identifier without requiring training. Appendix~\ref{app:scope} gives a compact interface-level comparison.

Benchmarks spanning high-resolution search, mathematical reasoning, real-world perception, and agentic tool use stress the need to coordinate fine-grained evidence with multi-step reasoning~\cite{hrbench,vstar,mmerealworld,wemath,mathvision,mathvista,mathverse,logicvista,dynamath,visulogic,mmstar,realworldqa,mmmu,agentx,tirbench,vtcbench,agenticmme}. Evidence-oriented evaluation further suggests that accuracy alone may hide whether the right evidence was used~\cite{beyondaccuracy}. We therefore complement aggregate accuracy with matched visual-state controls, complete paired transitions, promotion usage, context-scaling diagnostics, and backend-specific latency measurements.

\section{Methodology}
\label{sec:method}

\begin{figure}[t]
    \centering
    \includegraphics[width=\linewidth]{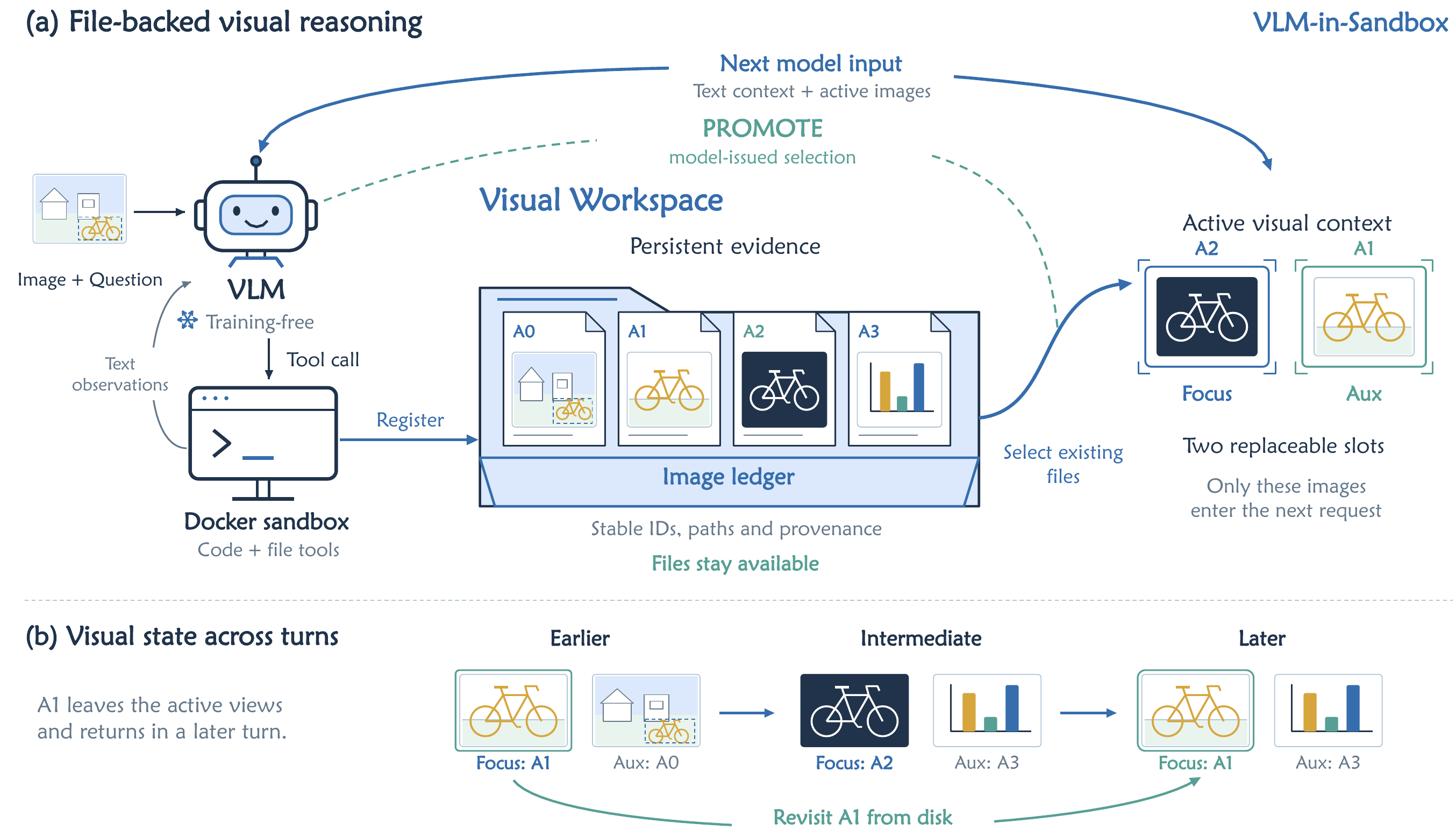}
    \caption{
    Overview of \textsc{VLM-in-Sandbox}. 
    (a) A frozen VLM uses harness-defined tools in a Docker sandbox to generate visual evidence. Visual Workspace preserves files with stable identifiers and provenance; model-issued \textsc{Promote} operations select the \texttt{focus} and \texttt{aux} images for the next request.
    (b) Replacing an active view leaves its file available for later selection. The illustrative, nonconsecutive snapshots show an earlier artifact returning to active context without replaying all intermediate images.
    }
    \label{fig:pipeline}
    \vspace{-4pt}
\end{figure}

We introduce \textsc{VLM-in-Sandbox}, a framework that equips VLMs with agentic multimodal reasoning in controlled computer environments.
Given an image set $I$ and a text query $q$, the agent interacts with a Docker sandbox through executable tools, produces intermediate visual artifacts, and writes a final answer to a designated file.
Our key design principle is the separation of \emph{visual evidence generation}, performed by sandbox tools, from \emph{visual evidence management}, handled by a runtime layer we call the \emph{Visual Workspace}.
Figure~\ref{fig:pipeline} shows the architecture, while Algorithm~\ref{alg:vlm-in-sandbox} in the appendix gives the full runtime loop; Figure~\ref{fig:motivation} (\S\ref{sec:introduction}) provides the worked HRBench-4K example we use throughout this section to ground each component in a concrete trace.

\subsection{Agent--Environment Loop}
\label{subsec:loop}

At each step $t$, the agent receives a multimodal prompt and either issues a tool call or produces a final answer.
Tool calls execute through a sandbox executor exposing three interfaces: \texttt{execute\_bash} for code execution, \texttt{str\_replace\_editor} for file manipulation and explicit promotion (\S\ref{subsec:active-context}), and \texttt{submit} to finalize the answer. Our implementation uses Docker to pin dependencies, isolate episodes, and clean up their files, but the executor is replaceable: the method requires the same file-and-tool contract rather than Docker-specific model behavior.
The sandbox serves as an external computation layer, supporting operations---cropping, zooming, analytic rendering, plot rendering, annotation, pixel-level statistics, diagnostic mask generation---that are difficult to perform reliably inside a single forward pass.
 
A defining property of multimodal sandboxing is that tool execution can produce new \emph{visual artifacts} that constitute intermediate evidence for later reasoning.
The central challenge is therefore not what tools the model can call, but how the resulting visual evidence is represented, indexed, and routed back into context.
In the \textbf{Append-only Sandbox} baseline, generated images are appended to the conversation and remain visible in all subsequent turns---a direct multimodal extension of the standard text-only sandbox paradigm.
This append-only scheme creates two compounding problems: redundant visual tokens accumulate as the trajectory grows and dilute attention, and the model cannot selectively revisit earlier evidence without it persisting through every intervening turn.
The Visual Workspace addresses both by decoupling artifact generation from artifact visibility.

\subsection{Image Ledger: Making Visual Evidence Addressable}
\label{subsec:ledger}
 
The \emph{image ledger} $\mathcal{L}_t$ is a structured registry for all visual artifacts known to the agent.
Each input or generated image receives a stable identifier, a file location, and provenance linking it to its source artifacts.
Stable identifiers turn anonymous sandbox files into addressable objects, which is essential for tasks requiring iterative localization: a model may first crop a broad region, then a sub-region, then zoom further, and reliable references across these chains are otherwise difficult to maintain.
 
\paragraph{Payload eviction.}
Each registered asset has an optional inline payload---the encoded image data used to render it as a visual input.
After every step we evict payloads from non-active, non-input assets (Algorithm~\ref{alg:vlm-in-sandbox}) while keeping their metadata in $\mathcal{L}_t$.
The model can still see that an asset exists and reference it by id, but its image data is reloaded from disk only when it is promoted back into active context.

Inactive image payloads can therefore be released without losing the evidence itself. Appendix~\ref{app:system} summarizes the workspace operations.

\subsection{Active Visual Context and Explicit Promotion}
\label{subsec:active-context}
 
\paragraph{Bounded active context.}
The \emph{active visual context} $\mathcal{C}_t$ is the subset of registered artifacts rendered as visual input in the next prompt.
We implement $\mathcal{C}_t$ as two named slots:
\begin{equation}
    \mathcal{C}_t \;=\; \bigl\{\,\texttt{focus}_t,\;\texttt{aux}_t\,\bigr\}.
    \label{eq:active-context}
\end{equation}
The \texttt{focus} slot holds the primary evidence the model currently inspects; \texttt{aux} holds a paired supporting view such as a parent crop or contextual region (e.g., focus on a zoomed-in object, aux on its surrounding scene).
Bounding the active context to two slots is deliberate: a sandboxed VLM can easily generate dozens of images, and re-injecting them all creates redundant visual tokens and expands every subsequent request.
The model can still browse the full ledger through textual indices, but only promoted artifacts appear as image inputs.

\paragraph{Explicit promotion.}
Generated artifacts are \emph{not} automatically visible in subsequent turns.
The model must issue an explicit promotion:
\begin{equation}
    \textsc{Promote}(a,\, s), \quad a \in \mathcal{L}_t,\; s \in \{\texttt{focus},\,\texttt{aux}\},
    \label{eq:promote}
\end{equation}
which moves asset $a$ from the ledger into slot $s$ of the active context.
Operationally, this is a subcommand of the editor interface (e.g., \texttt{promote asset://asset\_0008 slot=focus}); it performs no image processing, only routing.
This narrow semantics is important: in Append-only Sandbox, every crop enters history simply because it was produced; in the Visual Workspace, a crop becomes active evidence only when explicitly selected, separating \emph{producing} evidence from \emph{deciding which evidence guides future reasoning}.
This also makes negative or exploratory visual work cheap: the agent may create several candidate crops, masks, or overlays during search, while only the artifacts judged useful need to consume active visual slots.
Thus \textsc{Promote} is not an additional visual tool, but a routing decision over already-created evidence.

For example, the HRBench-4K agent generates several candidate color masks and registers them as visual artifacts. It promotes their combined visualization to \texttt{focus}, compares the evidence, and answers yellow. The remaining artifacts stay on disk without occupying active visual slots. A promotion changes visibility, not the underlying files:
\begin{quote}
\small\texttt{promote asset://masks\_combined slot=focus}
\end{quote}
The identifier above is illustrative; Figure~\ref{fig:motivation} shows the selection mechanism and the answer from the original run.

\subsection{Prompt Compilation}
\label{subsec:prompt-compilation}
 
Rather than replaying an unbounded multimodal history, the \emph{prompt compiler} deterministically reconstructs each model request from the workspace state:
\begin{equation}
    x_t \;=\;
    \textsc{Compile}\!\bigl(
        q,\; \mathcal{T}_{[-k:]},\; \textsc{Recap}(\mathcal{T}_{<\!-k}),\;
        \mathcal{L}_{t-1},\; \mathcal{C}_{t-1},\; g_t
    \bigr),
    \label{eq:compile}
\end{equation}
where $\mathcal{T}_{[-k:]}$ are the $k$ most recent uncompressed action--observation turns ($k{=}3$ by default), $\textsc{Recap}(\cdot)$ is a deterministic textual summary of older turns (action types and observation digests, no inline images), and $g_t$ is step-budget guidance discussed below.
The compiled prompt concatenates fixed system and task instructions, the older-turn recap, the recent-turns block, a workspace state summary listing asset metadata and recent-artifact indices, the deadline guidance, and the active visual context rendered as inline images.
Only artifacts in $\mathcal{C}_{t-1}$ are rendered visually in request $x_t$; all other registered assets are referenced only through textual metadata.
This prevents the failure mode of naive sandboxing---where every generated crop, rendered diagram, and diagnostic image accumulates in the prompt---and bounds visual context size independently of trajectory length.
 
The runtime also provides deterministic budget reminders and checks that an answer is available before accepting submission. These policies are shared by all matched configurations; Appendix~\ref{app:setup} gives their settings.

\begin{table}[!t]
\centering
\caption{%
\textbf{Main results across closed-source and open-source MLLMs.}
We compare three inference settings (Vanilla VLM, + Sandbox Only, and + Ours)
across seven multimodal benchmarks grouped into three task families:
high-resolution visual search (HRBench-4K, V$^{*}$Bench),
mathematical reasoning (WeMath, MathVision),
and general multimodal reasoning (MMStar, RealWorldQA, MMMU).
Each cell reports accuracy (\%).
Within each model block, the best accuracy is shown in \textbf{bold}.
The \textbf{Avg.} column reports a sample-size-weighted average across the seven benchmarks.
}
\label{tab:main_results_acc}

\setlength{\tabcolsep}{5pt}
\renewcommand{\arraystretch}{1.08}

\resizebox{\linewidth}{!}{%
\begin{tabular}{l l cc cc ccc c}
\toprule

\multicolumn{1}{c}{\multirow[c]{2}{*}{\textbf{Model}}}
&
\multicolumn{1}{c}{\multirow[c]{2}{*}{\textbf{Setting}}}
&
\multicolumn{2}{c}{
  \cellcolor{catVS}
  \textbf{\faSearch\;Search}
}
&
\multicolumn{2}{c}{
  \cellcolor{catMath}
  \textbf{\faCalculator\;Math}
}
&
\multicolumn{3}{c}{
  \cellcolor{catGen}
  \textbf{\faPuzzlePiece\;General}
}
&
\\


&
&
\cellcolor{catVS}\textbf{HRBench-4K}
&
\cellcolor{catVS}\textbf{V$^{*}$Bench}
&
\cellcolor{catMath}\textbf{WeMath}
&
\cellcolor{catMath}\textbf{MathVision}
&
\cellcolor{catGen}\textbf{MMStar}
&
\cellcolor{catGen}\textbf{RealWorldQA}
&
\cellcolor{catGen}\textbf{MMMU}
&
\multirow[c]{-2}{*}{\textbf{Avg.}}
\\

\midrule
\multicolumn{10}{c}{\textbf{Closed-source Models}} \\
\midrule

\multirow[c]{3}{*}{\textbf{GPT-4.1-mini}}
  & Vanilla VLM
  & 78.00 & 68.06 & 62.64 & 41.12
  & 64.40 & 71.76 & 66.19 & 65.81 \\

  & + Sandbox Only
  & 77.63 & 68.06 & 62.76 & 44.08
  & 60.47 & 70.33 & 66.86 & 64.95 \\

\rowcolor{tabtop1purple}
  \cellcolor{white} & + Ours
  & \bestval{79.25} & \bestval{70.68}
  & \bestval{64.08} & \bestval{46.71}
  & \bestval{65.67} & \bestval{72.94}
  & \bestval{68.10} & \bestval{67.47} \\

\midrule

\multirow[c]{3}{*}{\textbf{Gemini-2.5-Flash}}
  & Vanilla VLM
  & 83.50 & 79.06 & 81.38 & 39.14
  & 76.87 & 74.77 & 71.81 & 76.11 \\

  & + Sandbox Only
  & 83.25 & 81.15 & 85.46 & 61.18
  & 76.67 & 75.03 & 72.48 & 78.41 \\

\rowcolor{tabtop1purple}
  \cellcolor{white} & + Ours
  & \bestval{84.75} & \bestval{83.25}
  & \bestval{86.49} & \bestval{62.17}
  & \bestval{78.00} & \bestval{76.08}
  & \bestval{73.81} & \bestval{79.65} \\

\midrule
\multicolumn{10}{c}{\textbf{Open-source Models}} \\
\midrule

\multirow[c]{3}{*}{\textbf{Qwen3.5-9B}}
  & Vanilla VLM
  & 77.25 & 89.53 & 76.84 & 42.11
  & 74.40 & \bestval{75.03} & 65.81 & 72.99 \\

  & + Sandbox Only
  & 77.75 & 84.29 & 86.90 & 55.92
  & 73.87 & 70.59 & 72.67 & 76.79 \\

\rowcolor{tabtop1purple}
  \cellcolor{white} & + Ours
  & \bestval{78.13} & \bestval{91.10}
  & \bestval{87.93} & \bestval{57.89}
  & \bestval{75.00} & 71.90
  & \bestval{74.10} & \bestval{78.08} \\

\midrule

\multirow[c]{3}{*}{\textbf{Ministral-3-8B-Instruct}}
  & Vanilla VLM
  & 51.50 & 49.74 & 45.23 & 24.67
  & 49.73 & 55.69 & 51.33 & 48.50 \\

  & + Sandbox Only
  & 52.38 & 43.98 & 64.54 & 28.29
  & 57.40 & 56.08 & 57.52 & 56.79 \\

\rowcolor{tabtop1purple}
  \cellcolor{white} & + Ours
  & \bestval{53.50} & \bestval{51.83}
  & \bestval{65.80} & \bestval{29.61}
  & \bestval{58.33} & \bestval{57.25}
  & \bestval{58.57} & \bestval{58.11} \\

\bottomrule
\end{tabular}%
}
\end{table}

\section{Experimental Evaluation}
\label{sec:experiments}

We first evaluate the complete method across models and task families, then use matched controls to separate visual selection from bounded retention. Paired outcomes and repeated runs test the reliability of the gains. Section~\ref{sec:workspace-analysis} examines how these choices affect context growth, inference cost, and individual reasoning trajectories.

\subsection{Experimental Setup and Main Results}
\label{subsec:setup}
\label{subsec:main_results}

\textbf{Benchmarks.}
We use seven benchmarks in three task families: high-resolution visual search---HRBench-4K~\cite{hrbench} and V$^{*}$Bench~\cite{vstar}; multimodal mathematical reasoning---WeMath (\textit{testmini})~\cite{wemath} and MathVision-mini~\cite{mathvision}; and general multimodal reasoning---MMStar~\cite{mmstar}, RealWorldQA~\cite{realworldqa}, and MMMU (\textit{dev/val})~\cite{mmmu}. The complete collection contains 6,350 examples. We report per-benchmark accuracy and sample-size-weighted pooled accuracy.

\begin{figure}[t]
\centering
\includegraphics[width=\textwidth]{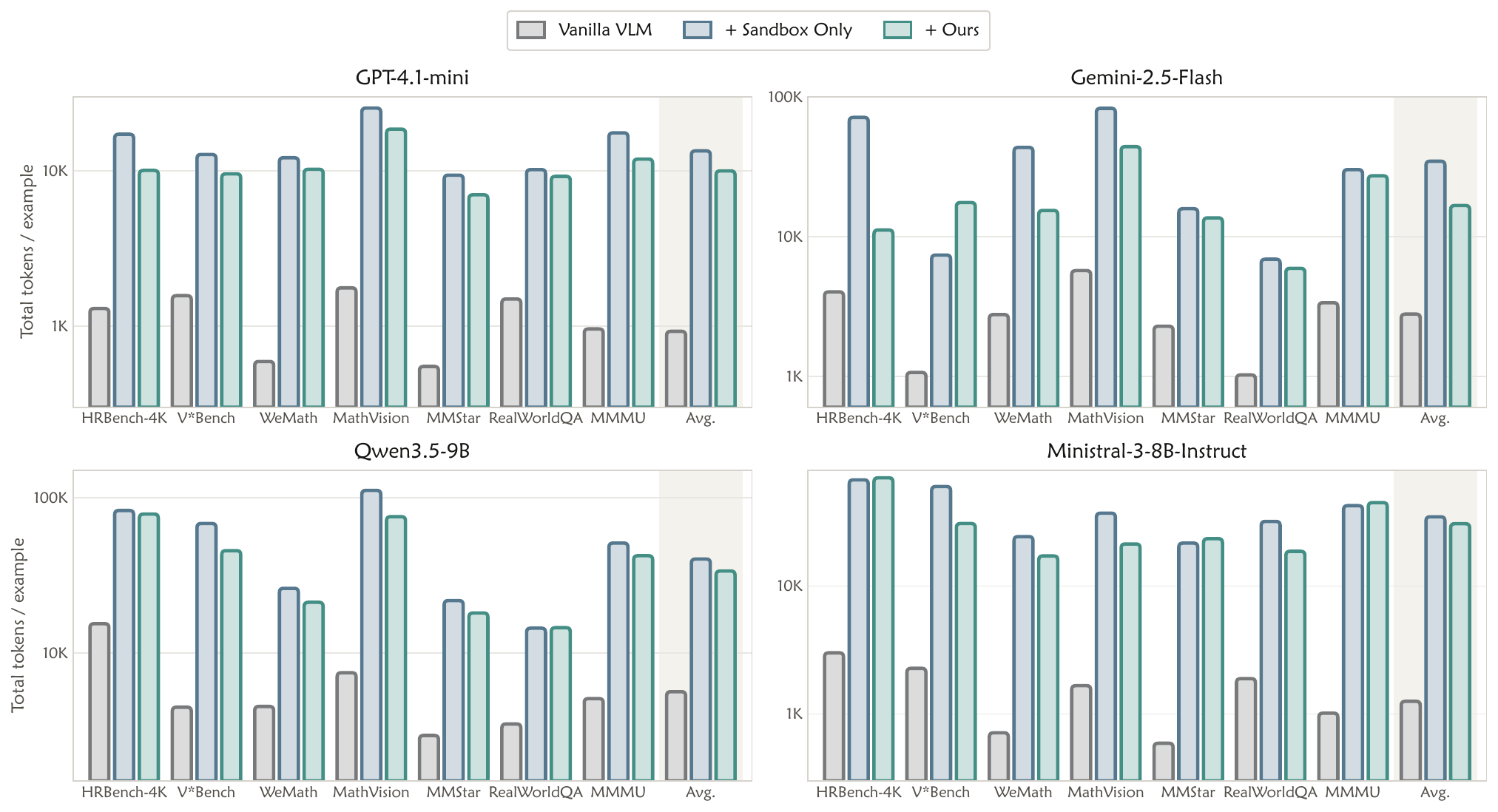}
\caption{\textbf{Average total-token usage across inference settings.} Bars report the average total tokens per instance for each benchmark on a logarithmic scale; the shaded \textbf{Avg.} column is the sample-size-weighted average over the seven benchmarks. Sandbox access costs roughly an order of magnitude more total tokens than a single Vanilla VLM call, and the Visual Workspace removes part of that overhead: at the aggregate level, \textbf{+ Ours} uses fewer total tokens than \textbf{+ Sandbox Only} for every base model while achieving higher accuracy.}
\label{fig:token_usage_broad}
\end{figure}

\textbf{Models and broad settings.}
The broad evaluation uses two API models, GPT-4.1-mini and Gemini-2.5-Flash, and two locally served models, Qwen3.5-9B and Ministral-3-8B-Instruct. \textbf{Vanilla VLM} receives only the task image(s) and question. \textbf{Original Append-only Sandbox}, denoted \textbf{+ Sandbox Only} in the submitted main table, provides the same executable tools as our method but automatically appends every generated image to subsequent multimodal history. \textbf{VLM-in-Sandbox} is the complete method, including the Visual Workspace. All settings use a temperature of 1.0; sandboxed settings use at most 20 agent steps.

\textbf{Controlled and systems studies.}
The matched study uses GPT-4.1-mini on complete V$^{*}$Bench, MathVision-mini, and RealWorldQA splits ($N=1{,}260$). Its four configurations share the model, prompt compiler, tool environment, inference budgets, and scoring protocol; only visual visibility and retention differ. The local systems study evaluates the same three splits with Qwen3.5-9B served by vLLM on $8\times$H20 with tensor parallelism 8 and request concurrency 1.

\textbf{Metrics and token scopes.}
The broad study reports provider- or server-reported \emph{total tokens} summed across calls, matching the submitted analysis in Figure~\ref{fig:token_usage_broad}; these values are compared only among settings for the same base model and backend. The matched study reports \emph{total tokens} (prompt plus completion). The local vLLM study additionally separates cached and uncached prompt tokens; its token and latency values are compared only within that backend. HRBench uses its official fixed GPT-4o-mini yes/no evaluator on every row; the remaining benchmarks use deterministic or official benchmark-specific extractors and answer checkers. Appendix~\ref{app:setup} documents splits, scoring, and inference settings.

Table~\ref{tab:main_results_acc} shows that \textsc{VLM-in-Sandbox} (\textbf{+ Ours}) achieves the highest sample-size-weighted accuracy for every base model. Relative to Vanilla VLM, it improves GPT-4.1-mini, Gemini-2.5-Flash, Qwen3.5-9B, and Ministral-3-8B-Instruct by 1.66, 3.54, 5.09, and 9.61 percentage points, respectively. The improvement is broad rather than driven by a single benchmark: \textbf{+ Ours} outperforms Vanilla VLM in 27 of the 28 model--benchmark pairs, with Qwen3.5-9B on RealWorldQA as the only exception. The gains span both proprietary and open-source model families, with the largest aggregate improvement observed for Ministral-3-8B-Instruct.

The comparison with \textbf{+ Sandbox Only} shows the gain of the complete \textsc{VLM-in-Sandbox} design over sandbox access alone. Sandbox access is not uniformly beneficial: for example, it reduces the GPT-4.1-mini average from 65.81 to 64.95, whereas \textbf{+ Ours} raises it to 67.47. More generally, \textbf{+ Ours} improves over \textbf{+ Sandbox Only} in all 28 model--benchmark pairs. The largest gains over Vanilla VLM cluster on WeMath and MathVision, where executable computation is particularly useful. For both open-source models, the largest gain over \textbf{+ Sandbox Only} occurs on V$^{*}$Bench: 6.81 points for Qwen3.5-9B and 7.85 points for Ministral-3-8B-Instruct. The compiler-matched controls in Section~\ref{subsec:mechanism} then isolate visual selection and bounded retention from the surrounding tool harness.

Figure~\ref{fig:token_usage_broad} makes the broad inference cost explicit. Both tool-enabled settings generate substantially more tokens than a single Vanilla VLM call, so the relevant efficiency comparison is with the tool-using \textbf{+ Sandbox Only} baseline. On the sample-size-weighted average, \textbf{+ Ours} reduces token usage by 25.7\%, 51.8\%, 16.3\%, and 11.7\% for GPT-4.1-mini, Gemini-2.5-Flash, Qwen3.5-9B, and Ministral-3-8B-Instruct, respectively, while improving accuracy. At the model-level aggregate, \textsc{VLM-in-Sandbox} therefore improves both accuracy and total-token efficiency relative to \textbf{+ Sandbox Only}. The magnitude of the reduction varies across benchmarks, so we report the aggregate as a summary rather than a uniform per-task effect.

\FloatBarrier
\subsection{Matched Component Ablations}
\label{subsec:mechanism}

The broad comparison evaluates complete inference settings, but does not by itself distinguish selection from retention. We therefore hold the prompt compiler and executable tools fixed and vary these two factors on the complete three-benchmark collection.

\begin{figure}[htbp]
\centering
\includegraphics[width=\linewidth]{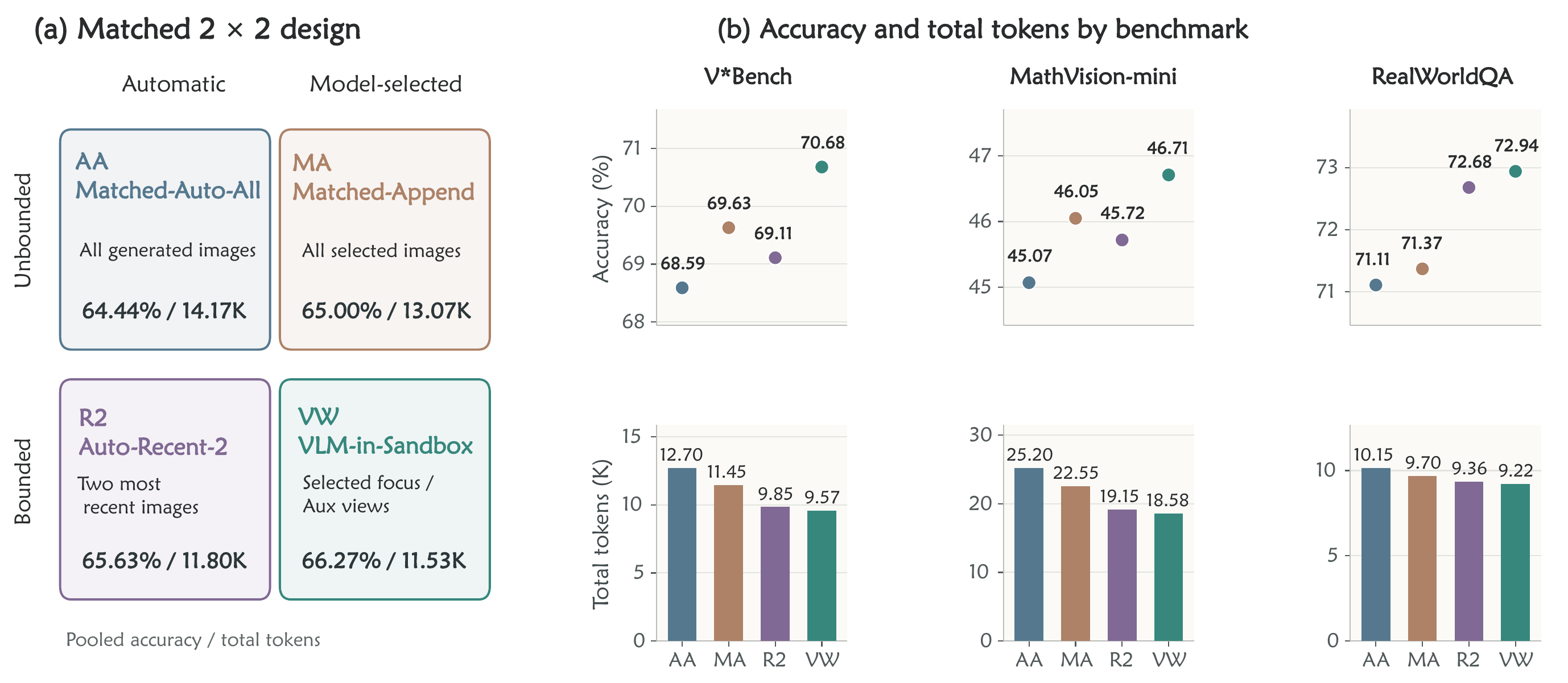}
\caption{\textbf{Matched visual-state controls.} Left: the $2\times2$ design, with sample-weighted accuracy and total tokens per example in each cell. Right: the four settings on complete V$^{*}$Bench ($N=191$), MathVision-mini ($N=304$), and RealWorldQA ($N=765$), using GPT-4.1-mini. Accuracy panels use separate, labeled scales; token bars start at zero. Original Append-only is not a factorial cell.}
\label{fig:matched_controls}
\end{figure}

Figure~\ref{fig:matched_controls} presents a $2\times2$ visibility-by-retention study. Matched-Auto-All is the compiler-matched append-only control: every generated image becomes visible and remains in later requests. Matched-Append requires model selection through \textsc{Promote} but retains all selected images. Auto-Recent-2 automatically exposes only the two most recent derived images. VLM-in-Sandbox combines model-directed selection with two replaceable \texttt{focus}/\texttt{aux} slots.

VLM-in-Sandbox improves over Matched-Auto-All by 1.83 points, with 52 rescues and 29 regressions (paired bootstrap 95\% CI [0.40, 3.25]; exact McNemar $p=0.014$), while using 18.6\% fewer total tokens. Bounded retention gives the largest token reduction: 16.7\% under automatic visibility and 11.8\% after model selection. At two-image capacity, model selection improves over recency by 0.63 points with 2.3\% fewer tokens, although the accuracy increment is not statistically conclusive (95\% CI [$-0.16$, 1.51]; $p=0.185$). Thus the complete policy performs best, while the factorial comparison identifies bounded retention as the larger token lever.

The three benchmarks expose different parts of this trade-off. On V$^{*}$Bench, selected but unbounded evidence is more accurate than automatic recency (69.63\% versus 69.11\%), but uses more tokens. On RealWorldQA, automatic recency is already close to VLM-in-Sandbox (72.68\% versus 72.94\%). The complete policy has the highest point estimate on all three, while the importance of selection varies by task. Figure~\ref{fig:matched_full} reports all five settings with accuracy and token cost plotted separately, including Original Append-only as a reference outside the factorial; Table~\ref{tab:matched_pairwise} reports the paired comparisons.

\FloatBarrier
\subsection{Reliability and Recovery}
\label{subsec:reliability}

\begin{figure}[htbp]
\centering
\includegraphics[width=\linewidth]{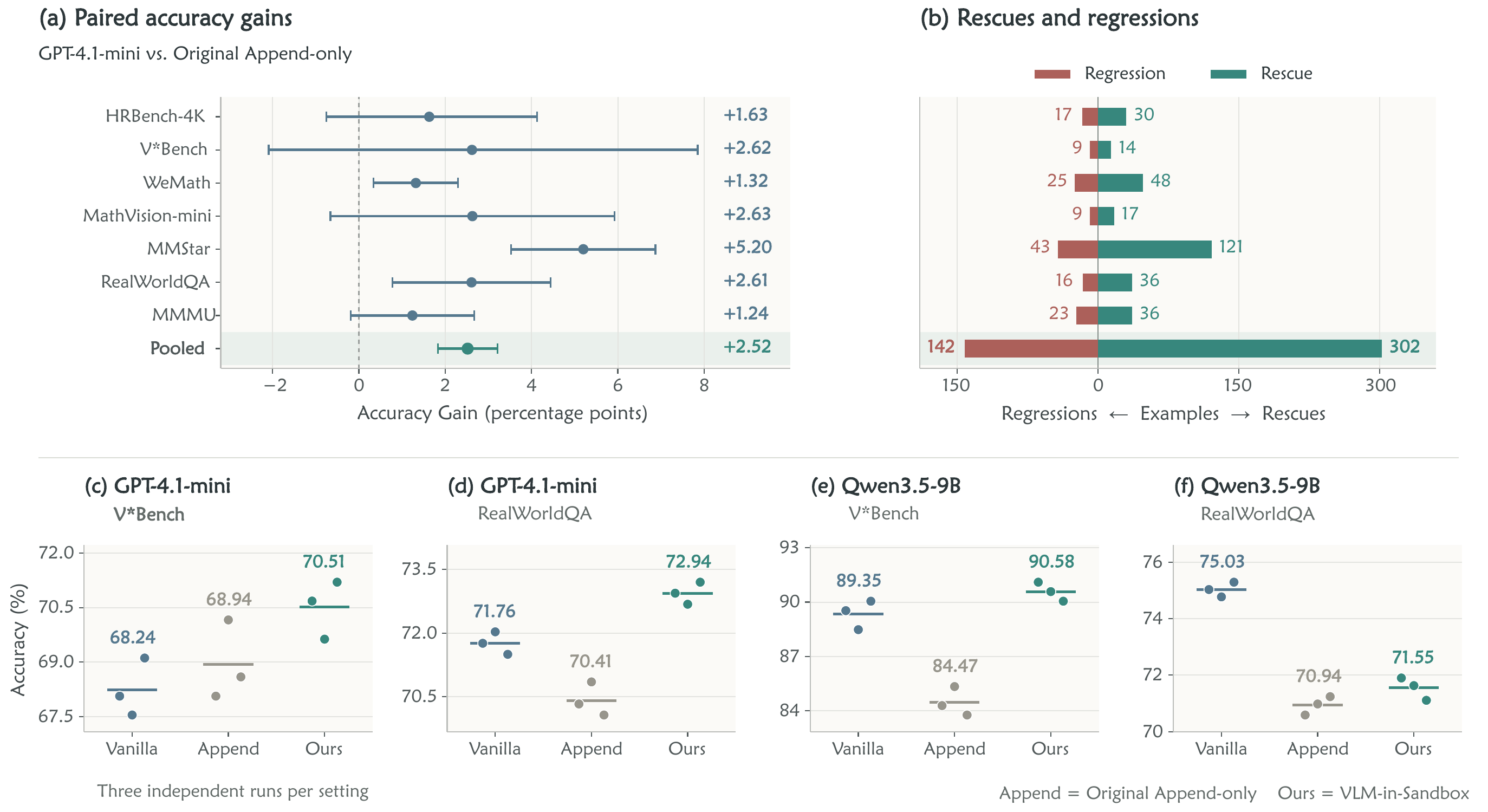}
\caption{\textbf{Reliability across examples and repeated runs.} (a--b) GPT-4.1-mini versus Original Append-only on seven benchmarks ($N=6{,}350$): paired accuracy gains with 95\% bootstrap intervals, and corresponding rescue/regression counts. Pooled resampling is benchmark-stratified with HRBench variants grouped by base question. (c--f) Three runs per setting at temperature 1.0 on complete V$^{*}$Bench and RealWorldQA splits. Points are individual runs; horizontal bars and labels show means. Accuracy scales differ across panels.}
\label{fig:paired_reliability}
\end{figure}

Figure~\ref{fig:paired_reliability} summarizes the paired results. Across all seven GPT-4.1-mini benchmarks, VLM-in-Sandbox converts 302 Original Append-only errors into correct answers while regressing on 142 previously correct examples: 2.13 rescues per regression and 160 additional correct answers. The resulting gain is 2.52 percentage points. The paired gain has a 95\% bootstrap CI of [1.83, 3.21] and $p<0.001$. Appendix~\ref{app:extended-results} specifies the stratified, question-clustered analysis and reports the complete per-benchmark results.

Across all four model--dataset combinations in Figure~\ref{fig:paired_reliability}(c--f), VLM-in-Sandbox has a higher three-run mean than Original Append-only. Repeated evaluation also preserves the main trade-off relative to Vanilla VLM. For Qwen3.5-9B, VLM-in-Sandbox has a higher mean on V$^{*}$Bench (90.58\% versus 89.35\%), but a lower mean on RealWorldQA (71.55\% versus 75.03\%). The complete seven-benchmark comparison yields 734 rescues and 411 regressions relative to Vanilla VLM. Thus sandbox reasoning solves many questions missed by the base model, but can also turn a correct direct answer into an error. Table~\ref{tab:complete_transitions} reports all four outcome transitions for both models, with their respective references.

\begin{table}[t]
\centering
\caption{\textbf{Wrong2Right analysis and promotion use by task family for Qwen3.5-9B.} Wrong2Right and Rate report the number and percentage of vanilla errors corrected by sandbox inference. Avg. Tool Calls reports the mean number of tool invocations per recovered example, and With Promote is the percentage invoking \texttt{promote} at least once.}
\label{tab:recovery_promotion}
\setlength{\tabcolsep}{8pt}
\renewcommand{\arraystretch}{1.10}
\begin{tabular}{cccccc}
\toprule
\textbf{Task} & \makecell{\textbf{Wrong} \xmark} & \makecell{\textbf{Wrong2Right} \recovermark} & \makecell{\textbf{Rate}} & \makecell{\textbf{Avg. Tool Calls}} & \makecell{\textbf{With Promote}} \\
\midrule
\cellcolor{catVS}\textbf{\faSearch\;Search} & 202 & 70 & 34.7\% & 5.64 & 27.1\% \\
\cellcolor{catMath}\textbf{\faCalculator\;Math} & 579 & 331 & 57.2\% & 6.15 & 4.2\% \\
\cellcolor{catGen}\textbf{\faPuzzlePiece\;General} & 934 & 333 & 35.7\% & 5.49 & 6.3\% \\
\midrule
\textbf{Overall} & \textbf{1,715} & \textbf{734} & \textbf{42.8\%} & \textbf{5.80} & \textbf{7.4\%} \\
\bottomrule
\end{tabular}
\end{table}

\paragraph{Wrong2Right Analysis and Selective Promotion.}

Table~\ref{tab:recovery_promotion} audits Qwen3.5-9B on all 1,715 examples missed by Vanilla VLM. \textsc{VLM-in-Sandbox} corrects 734 of them (42.8\%), with substantial variation across task families. Math achieves the highest recovery rate (57.2\%) and uses the most tools per recovered example (6.15), yet invokes \texttt{promote} in only 4.2\% of recoveries. This profile is consistent with computation-heavy trajectories whose symbolic or textual outputs can be consumed without visual re-entry. Search and General offer an informative contrast: their recovery rates (34.7\% versus 35.7\%) and average tool usage (5.64 versus 5.49 calls) are similar, while Search invokes \texttt{promote} 4.3$\times$ as often (27.1\% versus 6.3\%). Given their similar average tool counts, this pattern suggests that promotion frequency tracks the modality of intermediate evidence more closely than tool-call volume.

The recovery rate and promotion rate describe different quantities. The former measures how often \textsc{VLM-in-Sandbox} corrects a Vanilla VLM error; the latter measures how often a recovered trajectory invokes this routing operation. Recovered examples use 5.80 tool calls on average, but only 7.4\% invoke \texttt{promote}. Promotion is therefore selective rather than mandatory: many recoveries use textual tool outputs and persistent workspace state, while high-resolution search more often returns a crop, mask, or overlay to multimodal context. More broadly, the Visual Workspace keeps intermediate artifacts addressable and bounds the active visual context; \texttt{promote} provides targeted visual re-entry when needed. These conditional statistics characterize successful trajectories, while component attribution is established by the matched comparisons above.

\FloatBarrier

\section{Visual Workspace Analysis}
\label{sec:workspace-analysis}

The matched study establishes the accuracy--cost trade-off of visual-state policies. We now examine how a growing artifact collection translates into model-visible context, whether the smaller workload remains beneficial with prefix caching, and how selected evidence supports individual solutions.

\subsection{Context Growth and Visibility Control}
\label{subsec:context-growth}

Persistent storage and active visibility serve different purposes: stored files preserve evidence for later access, whereas active slots determine which derived images enter the next request. We test this distinction with fixed-trace replays and then examine the capacity and history settings on actual benchmark runs.

\begin{figure}[H]
\centering
\includegraphics[width=\linewidth]{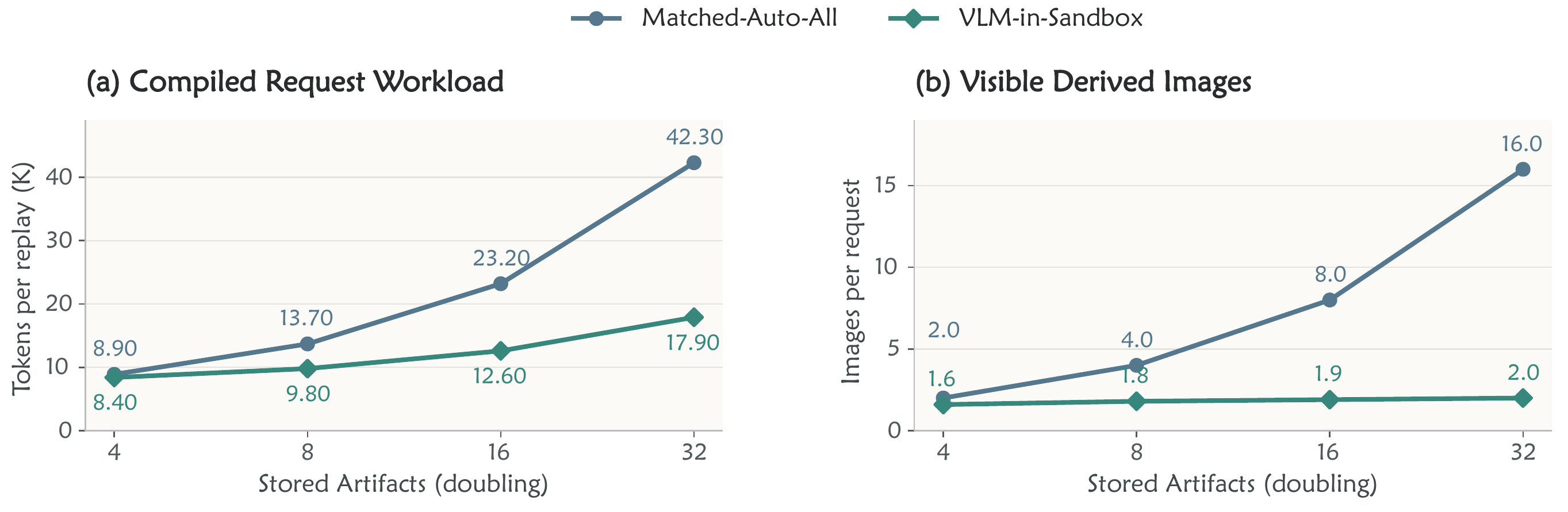}
\caption{\textbf{Stored artifacts versus model-visible context.} Open-loop replays of 100 successful trajectories, with actions, observations, selections, and termination fixed. Pixel-identical copies increase the artifact count without adding visual information. Left: tokens summed over compiled requests per replay. Right: mean visible derived images per request, excluding the source image. No model calls are made.}
\label{fig:context_growth}
\end{figure}

Figure~\ref{fig:context_growth} increases stored artifacts to 4, 8, 16, and 32. Matched-Auto-All grows from 8.90K to 42.30K tokens per replay, whereas VLM-in-Sandbox grows from 8.40K to 17.90K while keeping visible derived images near two. The workspace therefore decouples the size of the stored collection from the number of images repeatedly exposed to the model. Its total token count still grows with textual history and artifact metadata; bounded visual slots do not imply constant total context. This diagnostic measures request scaling, not task accuracy or closed-loop long-horizon performance. Appendix~\ref{app:extended-results} specifies the replay construction.

\begin{figure}[H]
\centering
\includegraphics[width=0.88\linewidth]{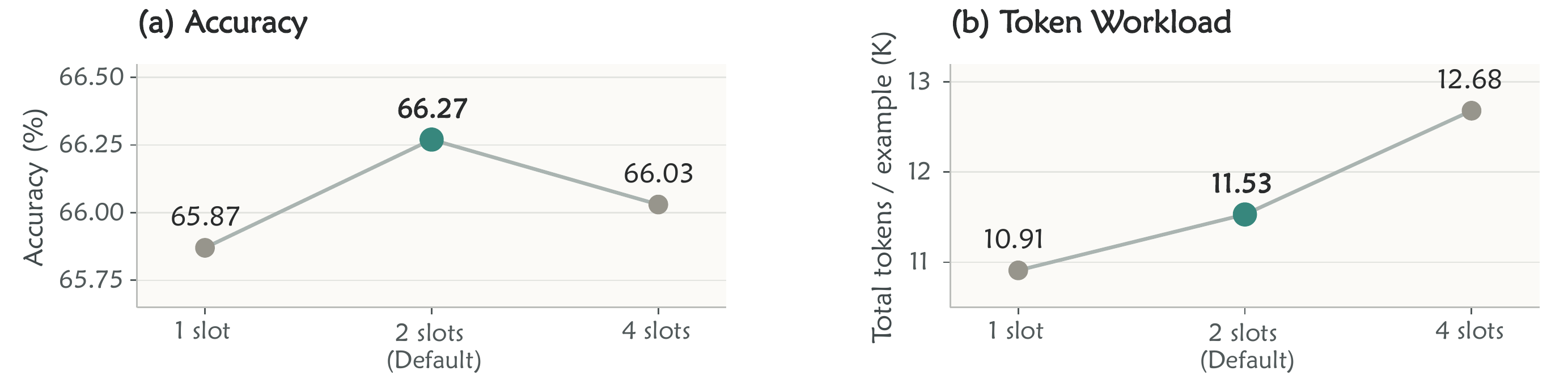}
\caption{\textbf{Active visual capacity.} Sample-weighted GPT-4.1-mini results on the matched-study collection ($N=1{,}260$). Separate panels show accuracy and mean total tokens per example; the two-slot default is highlighted. Differences are descriptive point estimates.}
\label{fig:slot_capacity}
\end{figure}

Figure~\ref{fig:slot_capacity} examines whether a small active context is sufficient. One slot is cheapest, two slots achieve the highest observed accuracy, and four slots increase token use without improving the point estimate. These results support a compact working set, without establishing a statistically unique optimum. The ledger retains files regardless of slot replacement, so a smaller active context need not discard the underlying evidence.

Textual history contributes separately to the request workload. Replacing unlimited raw replay with three recent turns plus deterministic recap reduces total tokens from 12.61K to 11.53K (8.6\%) and changes accuracy from 65.87\% to 66.27\%. This comparison changes both the replay window and older-turn recap; it evaluates the combined history policy rather than recap alone. Table~\ref{tab:step_deadline} complements the context analysis with step-budget sensitivity.

\FloatBarrier
\subsection{Inference Efficiency}
\label{subsec:inference-efficiency}

A smaller request does not automatically imply a proportional latency reduction: prefix caching can reuse previously processed inputs, and model, vision, and tool execution all contribute to runtime. We therefore compare all five settings under one local serving configuration.

\begin{figure}[H]
\centering
\includegraphics[width=\linewidth]{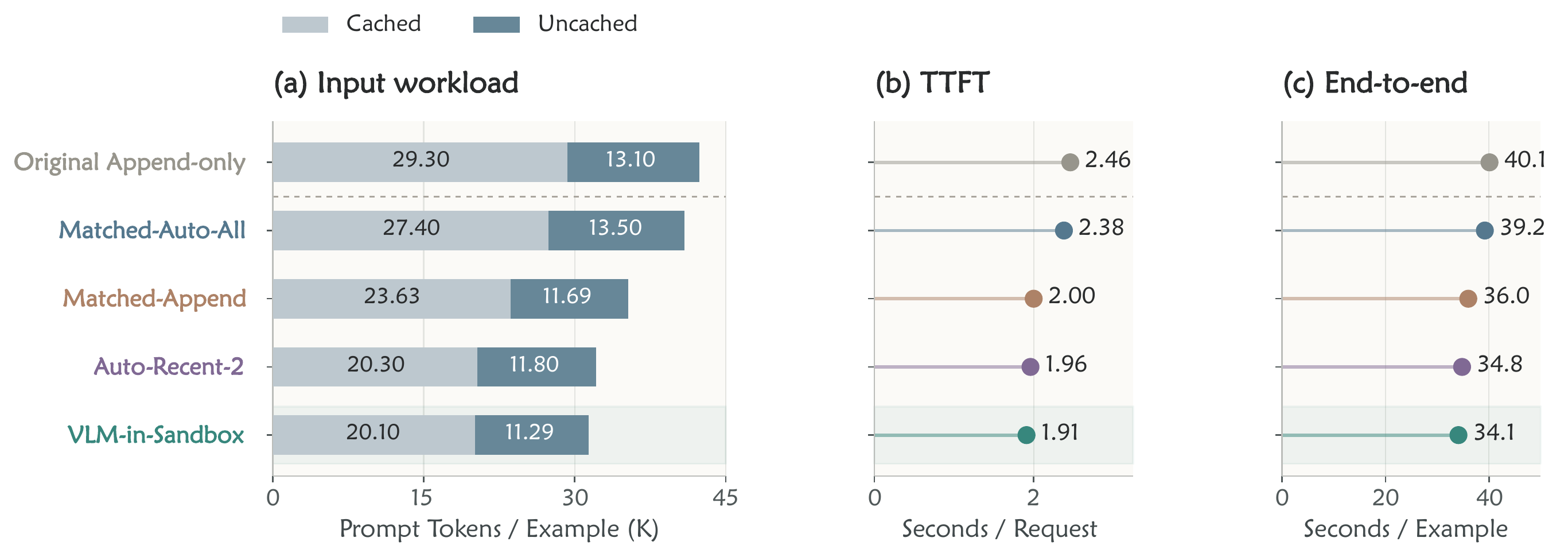}
\caption{\textbf{Inference with prefix caching.} Qwen3.5-9B served by vLLM on $8\times$H20 (tensor parallelism 8, request concurrency 1), on the same three complete benchmark splits ($N=1{,}260$). Stacked bars separate cached and uncached prompt tokens per example. TTFT is measured per request; E2E is measured per example. The dashed divider separates Original Append-only from the matched configurations.}
\label{fig:cache_efficiency}
\end{figure}

Figure~\ref{fig:cache_efficiency} shows that prefix reuse does not reverse the efficiency advantage. VLM-in-Sandbox has a lower cache-hit ratio than Matched-Auto-All (64.0\% versus 67.0\%), but also less uncached input (11.29K versus 13.50K tokens per example). Relative to that control, it reduces total tokens by 23.5\%, uncached prompt tokens by 16.4\%, TTFT by 19.7\%, and E2E by 13.0\%. The lower hit ratio reflects a different request workload; it is not, by itself, evidence of a higher inference cost.

Each configuration starts from an empty cache. Timing includes multimodal preprocessing and vision encoding; E2E additionally includes sandbox lifecycle and excludes offline scoring. These measurements characterize this local backend, not API pricing. Table~\ref{tab:cache_full} gives the complete prompt, completion, cache, and latency breakdown.

Container creation, startup, and removal take 0.82 seconds per example in the local $N=1{,}260$ evaluation, or 2.0\% of Original Append-only's 40.1-second E2E time. No create/start/remove or resource-limit failures are observed. This is measured lifecycle time, not a host-versus-container difference; Docker remains the execution backend rather than an accuracy-enhancing component. Appendix~\ref{app:docker} reports the resource configuration and failure accounting.

\FloatBarrier
\subsection{Qualitative Analysis}
\label{subsec:qualitative}

\begin{figure}[H]
    \centering
    \includegraphics[width=\linewidth]{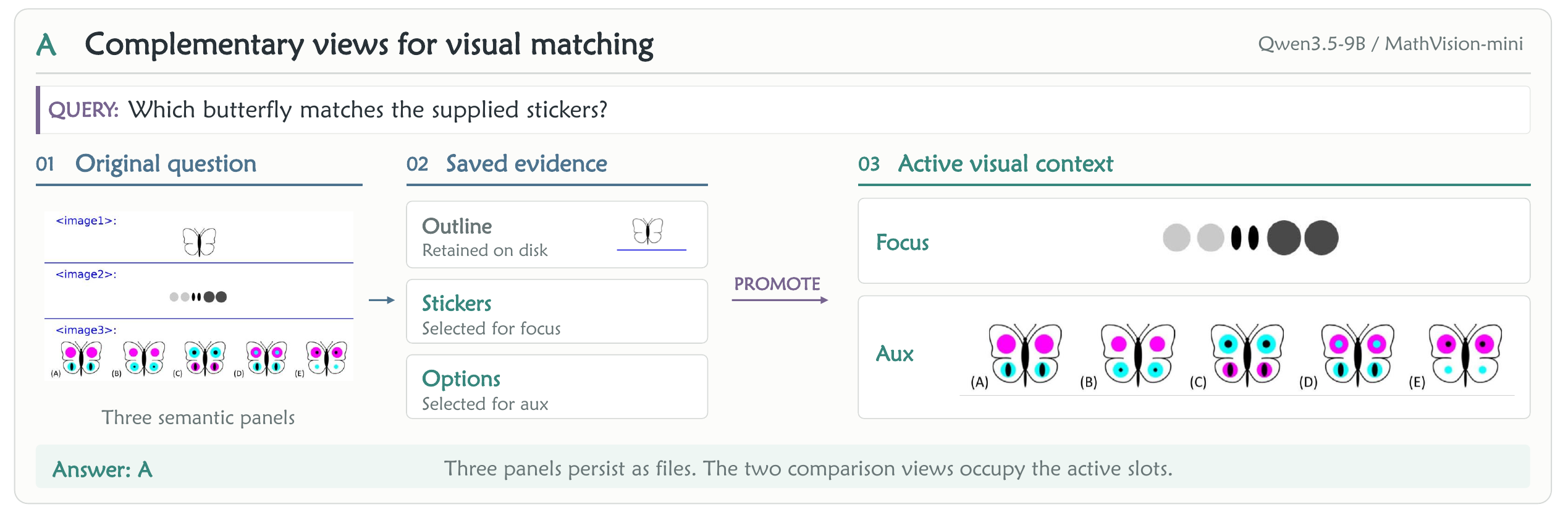}
    \caption{\textbf{Complementary views for visual matching} (Qwen3.5-9B, MathVision-mini). The agent retains three panels as files and selects the stickers and candidates for the two active slots. Their comparison supports answer A.}
    \label{fig:case_study_butterfly}
\end{figure}

\begin{figure}[H]
    \centering
    \includegraphics[width=\linewidth]{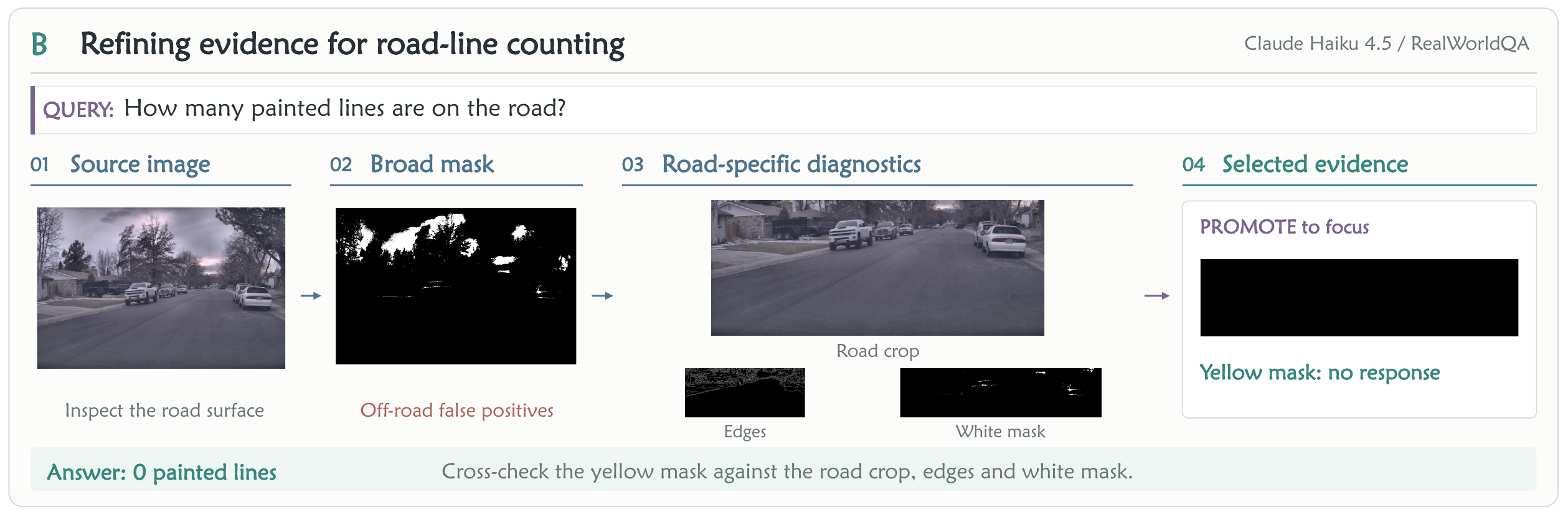}
    \caption{\textbf{Refining evidence for road-line counting} (Claude Haiku 4.5, RealWorldQA). Selected stages of a longer trajectory show a broad mask followed by road-specific diagnostics. The promoted yellow mask is considered together with the crop, edges, and white mask before the answer of zero painted lines.}
    \label{fig:case_study_roadline}
\end{figure}

\begin{figure}[H]
    \centering
    \includegraphics[width=\linewidth]{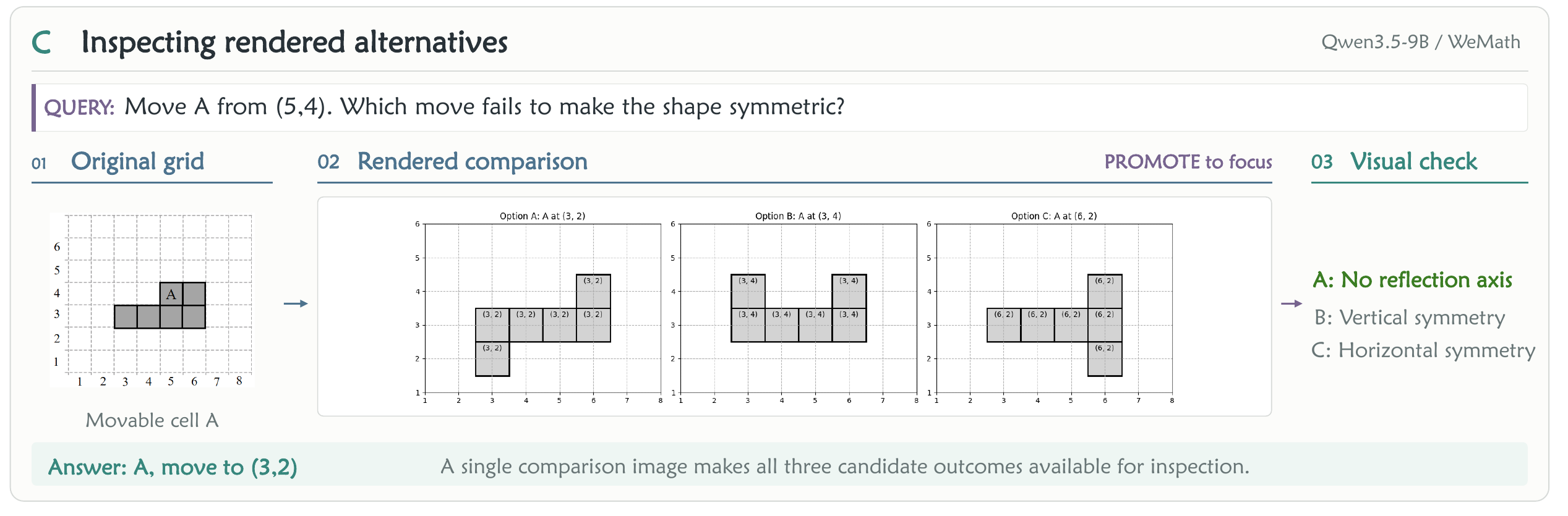}
    \caption{\textbf{Inspecting rendered alternatives} (Qwen3.5-9B, WeMath). The agent renders the three candidate moves and promotes the comparison image to \texttt{focus}. Moving A to $(3,2)$ fails to produce reflection symmetry; the other two moves succeed. The original generated rendering is preserved.}
    \label{fig:case_study_symmetry}
\end{figure}

Figures~\ref{fig:case_study_butterfly}--\ref{fig:case_study_symmetry} show three uses of selected visual evidence: comparing complementary views, refining diagnostics, and inspecting rendered alternatives. The cases illustrate reasoning behavior rather than replacing the matched comparisons in Section~\ref{subsec:mechanism}.

In the butterfly puzzle, the agent separates a composite into outline, sticker, and candidate panels, then selects the sticker panel as \texttt{focus} and the candidates as \texttt{aux}. The selected views support direct comparison while the remaining panel stays available as a file.

In the road-line example, a broad bright-pixel mask responds to trees and vehicles. The agent narrows its analysis to road crops, edges, and color masks, including a promoted yellow mask, before answering zero. The empty yellow mask is interpreted alongside the other views, not as standalone proof that markings are absent.

In the symmetry example, the agent renders the consequences of three candidate moves in a single comparison image and promotes it to identify the move that does not preserve a reflection axis. The butterfly and symmetry examples use Qwen3.5-9B; the road example is an additional Claude Haiku 4.5 trajectory, separate from the four-model main comparison.

These examples share a common pattern: tools produce inspectable evidence, and the model chooses which evidence to bring back into context. Selection alone does not guarantee a correct interpretation. Appendix~\ref{app:evidence_failure} shows an overlay marking vehicle rears that is misread as pavement arrows despite a correct final answer, illustrating why evidence quality and final-answer accuracy must remain distinct.

\FloatBarrier

\section{Discussion and Limitations}
\label{sec:discussion}

\textbf{Computation and deployment.}
Tool-enabled reasoning costs more than a single Vanilla VLM call. Visual Workspace reduces the repeated visual workload relative to append-only sandboxing, but does not eliminate model and tool execution costs. Docker supplies a reproducible filesystem and execution environment; it can be replaced by a backend with equivalent tool behavior. Deployments still require resources and appropriate execution permissions.

\textbf{Model dependence.}
Because the framework is training-free, the base VLM must decide when tools are useful, how long to continue, and which evidence deserves visual re-entry. This creates a real trade-off: Qwen3.5-9B obtains 734 rescues but also 411 regressions relative to Vanilla VLM, and its negative RealWorldQA cell is associated with longer, artifact-heavy trajectories on questions often answerable directly. Correctness-independent routing that selectively invokes sandbox reasoning is therefore an important extension.

\textbf{Scope and extension to video.}
We evaluate static-image reasoning. Persistent visual files and model-directed selection could extend to video by retaining temporal evidence while exposing only relevant frames or clips in each request. Temporal retrieval and tracking remain future work and require dedicated evaluation.

\textbf{Responsible use.}
Tool-enabled VLMs may process sensitive images or be over-trusted in high-stakes settings. Deployment should provide appropriate access controls and human oversight.

\section{Conclusion}

We introduced \textsc{VLM-in-Sandbox}, a training-free framework that separates persistent visual evidence from the images exposed to the model. Its Visual Workspace combines addressable artifacts, bounded active slots, and model-directed promotion. Across seven benchmarks and four VLMs, the complete method improves sample-weighted accuracy over Vanilla VLM and Original Append-only Sandbox. Matched controls identify bounded retention as the larger source of token savings, while selected visibility achieves the best observed accuracy. Paired outcomes, scaling replays, and prefix-cached measurements characterize the gains and their trade-offs. These findings support explicit visual evidence management as a core abstraction for sandboxed VLM reasoning.

\FloatBarrier
\clearpage

\bibliographystyle{plainnat}
\bibliography{main}

\FloatBarrier
\clearpage
\appendix
\section{System and Reproducibility}
\label{app:system}

\subsection{Runtime Interface and Algorithm}

The Visual Workspace is a typed visual-artifact state layer. Its minimal operations are summarized in Table~\ref{tab:vw_interface}; model history, step budgets, deadline guidance, answer handling, and the Docker executor are integration policies surrounding this interface.

\begin{table}[H]
\centering
\caption{\textbf{Minimal Visual Workspace interface.} The interface is independent of the concrete sandbox executor.}
\label{tab:vw_interface}
\begin{normaltable}
\begin{tabularx}{\linewidth}{L{0.23\linewidth}Y}
\toprule
\tblheaderrow
\tblhead{Operation} & \tblhead{State transition or returned information} \\
\midrule
{\sffamily\bfseries Register}$(f,\tau,p,P)$ & Register file $f$ with type $\tau$, provenance $p$, and optional parents $P$; return a stable asset identifier. \\
{\sffamily\bfseries Resolve/List} & Resolve an identifier to metadata or enumerate registered artifacts using metadata filters. \\
{\sffamily\bfseries Promote}$(a,s)$ & Put asset $a$ into active slot $s\in\{\texttt{focus},\texttt{aux}\}$ and return a promotion receipt. \\
{\sffamily\bfseries ActiveSlots} & Return the identifiers and metadata currently rendered as visual inputs. \\
{\sffamily\bfseries EvictPayload}$(a)$ & Release non-active inline image data while preserving the file, identifier, and metadata. \\
\bottomrule
\end{tabularx}
\end{normaltable}
\end{table}

Algorithm~\ref{alg:vlm-in-sandbox} expands the runtime loop. The first source image initially seeds \texttt{focus} and remains registered in the ledger. All compiler-matched settings use the same source-image initialization; reported derived-image counts exclude this source. Replacing \texttt{focus} changes which asset is active but does not delete the source file or its ledger entry.

\begin{algorithm}[H]
\caption{VLM-in-Sandbox with Visual Workspace}
\label{alg:vlm-in-sandbox}
\begin{algorithmic}[1]
\Require Images $I$, query $q$, maximum steps $T_{\max}$, raw-turn window $k$, deadline horizon $d$
\State $\mathcal{L}_0 \gets \textsc{Register}(I)$
\State $\mathcal{C}_0 \gets \{\texttt{focus}:I_1,\ \texttt{aux}:\varnothing\}$; $\mathcal{T}\gets\varnothing$
\For{$t=1,\ldots,T_{\max}$}
    \State $r_t\gets T_{\max}-t+1$
    \State $g_t\gets\textsc{DeadlineGuidance}(r_t,d,\textsc{AnswerFileExists}())$
    \State $x_t\gets\textsc{Compile}(q,\mathcal{T}_{[-k:]},\textsc{Recap}(\mathcal{T}_{<-k}),\mathcal{L}_{t-1},\mathcal{C}_{t-1},g_t)$
    \State $a_t\gets\textsc{VLM}(x_t)$
    \If{$a_t=\textsc{Submit}$ and \textsc{AnswerFileExists}()}
        \State \Return designated answer-file contents
    \ElsIf{$a_t=\textsc{Submit}$}
        \State $(\mathcal{L}_t,\mathcal{C}_t)\gets(\mathcal{L}_{t-1},\mathcal{C}_{t-1})$
        \State $o_t\gets$ missing-answer-file rejection
    \ElsIf{$a_t=\textsc{Promote}(a,s)$ and $a\in\mathcal{L}_{t-1}$}
        \State $\mathcal{L}_t\gets\mathcal{L}_{t-1}$; $\mathcal{C}_t\gets\mathcal{C}_{t-1}[s\mapsto a]$
        \State $o_t\gets$ promotion receipt
    \Else
        \State $(o_t,\Delta\mathcal{L})\gets\textsc{Execute}(a_t)$
        \State $\mathcal{L}_t\gets\mathcal{L}_{t-1}\cup\Delta\mathcal{L}$; $\mathcal{C}_t\gets\mathcal{C}_{t-1}$
    \EndIf
    \State $\mathcal{L}_t\gets\textsc{EvictPayloads}(\mathcal{L}_t,\textsc{Assets}(\mathcal{C}_t)\cup I)$
    \State append $(a_t,o_t)$ to $\mathcal{T}$
\EndFor
\State \Return answer-file contents if present; otherwise no answer
\end{algorithmic}
\end{algorithm}

Legacy image reads are normalized to \texttt{promote} and counted as promotion.

\subsection{Prompts, Tools, and Evaluation Settings}
\label{app:setup}

The compiler reconstructs every request from canonical state rather than mutating an unbounded chat transcript. Table~\ref{tab:prompt_contract} specifies the request blocks used by VLM-in-Sandbox and the matched controls.

\begin{table}[H]
\centering
\caption{\textbf{Compiled request structure.} Blocks appear in the listed logical order; active images are rendered as multimodal parts rather than text paths.}
\label{tab:prompt_contract}
\begin{prosetable}
\begin{tabularx}{\linewidth}{L{0.22\linewidth}Y}
\toprule
\tblheaderrow
\tblhead{Block} & \tblhead{Contents} \\
\midrule
{\sffamily\bfseries Fixed task block} & System instructions, benchmark question, output location, and tool-use contract. \\
{\sffamily\bfseries Older-turn recap} & Deterministic summaries of earlier actions, observations, and relevant artifact references. \\
{\sffamily\bfseries Recent history} & The three most recent assistant/action/observation turns replayed without compression in the submitted and matched configurations. \\
{\sffamily\bfseries Workspace summary} & Current active slots and a compact index of available visual artifacts. \\
{\sffamily\bfseries Runtime guidance} & Remaining-step count, answer-file presence, and deterministic deadline instruction. \\
{\sffamily\bfseries Active visual input} & Image payloads for the current \texttt{focus} and effective \texttt{aux} assets. \\
\bottomrule
\end{tabularx}
\end{prosetable}
\end{table}

\paragraph{System-prompt structure.}
All settings use common task instructions, file-based execution, and an answer-submission workflow. The excerpt below summarizes these shared instructions; it is not a verbatim full prompt.

\begin{lstlisting}[style=promptlisting]
Solve the task using the supplied images and sandbox tools when useful.
Save scripts and reusable visual artifacts in the workspace.
Base the answer on visual evidence and tool results.
Write only the final answer in the required format, then submit.
\end{lstlisting}

The visual-policy block distinguishes the settings. Original Append-only attaches generated images to conversation history. Within the matched comparison, Matched-Auto-All retains all generated images; Auto-Recent-2 retains the two most recent; Matched-Append retains all model-selected images; and VLM-in-Sandbox replaces the selected active slots. The matched configurations otherwise share their prompts and execution settings.

The model can execute programs, manipulate files, and submit answers through the interfaces in Table~\ref{tab:tool_schema}. Promotion routes an existing image; it does not perform image processing.

\begin{table}[H]
\centering
\caption{\textbf{Model-visible tool interfaces.}}
\label{tab:tool_schema}
\begin{normaltable}
\begin{tabularx}{\linewidth}{L{0.25\linewidth}Y}
\toprule
\tblheaderrow
\tblhead{Interface} & \tblhead{Function} \\
\midrule
\texttt{execute\_bash} & Execute code or shell commands and return text observations. \\
\texttt{str\_replace\_editor} & Read, create, and edit files; its \texttt{promote} command selects an existing image for an active slot. \\
\texttt{submit} & Finalize the answer stored in the designated output file. \\
\bottomrule
\end{tabularx}
\end{normaltable}
\end{table}

\paragraph{History and termination.}
Earlier turns are summarized deterministically from recorded actions and observations, without an additional model call. Budget reminders begin with three calls remaining: finish verification, write the answer, and submit. Submission requires an answer file; at the step limit, an existing non-empty answer is still scored. All matched configurations use these same rules.

Table~\ref{tab:core_settings} lists the common inference configuration. The matched controls change only automatic visibility and retention policy as specified in Figure~\ref{fig:matched_controls}; Original Append-only is a submitted reference that uses its original append-only history rather than a cell of the matched factorial.

\begin{table}[H]
\centering
\caption{\textbf{Core inference settings.}}
\label{tab:core_settings}
\begin{normaltable}
\begin{tabularx}{0.92\linewidth}{L{0.41\linewidth}Y}
\toprule
\tblheaderrow
\tblhead{Setting} & \tblhead{Value} \\
\midrule
Maximum agent steps $T_{\max}$ & 20 \\
Deadline horizon $d$ & 3 \\
Submitted/matched recent-turn window & Three turns plus deterministic recap \\
Active visual slots & \texttt{focus}, \texttt{aux} \\
Temperature & 1.0 \\
Maximum token limit & 131,072 in submitted and matched runs \\
Maximum generation request per call & 32,768 tokens \\
Maximum retained response length & 8,192 tokens \\
Request/tool timeout & 1,200 s per model request; 120 s per sandbox command \\
Sandbox backend & Docker executor \\
Local cache-study serving & vLLM on $8\times$H20, tensor parallelism 8 \\
\bottomrule
\end{tabularx}
\end{normaltable}
\end{table}

\paragraph{Scoring.}
We follow each benchmark's official evaluation protocol. HRBench-4K uses the same GPT-4o-mini yes/no judge across settings. V$^{*}$Bench, WeMath, MMStar, RealWorldQA, and MMMU use option parsing or normalized exact matching; MathVision-mini uses official answer extraction and symbolic/numeric equivalence. Multiple-choice responses contain the option letter; other questions require a short answer. Invalid or missing answers receive zero.

\subsection{Datasets and Experimental Configuration}

\begin{table}[H]
\centering
\caption{\textbf{Datasets and scoring protocols.} Public source identifiers and split names specify the evaluated releases. License statements follow the corresponding official dataset cards; mixed-source benchmarks remain subject to their upstream material terms.}
\label{tab:data_protocol}
\begin{normaltable}
\begin{tabularx}{\linewidth}{L{0.18\linewidth}L{0.39\linewidth}R{0.08\linewidth}Y}
\toprule
\tblgrouprow
\multicolumn{4}{c}{\tblhead{Evaluation splits and scoring}} \\
\midrule
\tblheaderrow
\tblhead{Dataset} & \tblhead{Split/release} & \tblhead{$N$} & \tblhead{Scorer} \\
\midrule
{\sffamily\bfseries HRBench-4K} & \texttt{hr\_bench\_4k.tsv}; 200 questions $\times$ four cyclic option orders & 800 & Official LLM judge \\
{\sffamily\bfseries V$^{*}$Bench} & \texttt{test} & 191 & Deterministic \\
{\sffamily\bfseries WeMath} & \texttt{testmini} & 1,740 & Deterministic \\
{\sffamily\bfseries MathVision-mini} & \texttt{testmini} & 304 & Official equivalence \\
{\sffamily\bfseries MMStar} & \texttt{val} & 1,500 & Deterministic \\
{\sffamily\bfseries RealWorldQA} & \texttt{test} & 765 & Deterministic \\
{\sffamily\bfseries MMMU} & Merged \texttt{dev}+\texttt{validation}, 30 subjects & 1,050 & Deterministic \\
\bottomrule
\end{tabularx}
\medskip
\begin{tabularx}{\linewidth}{L{0.18\linewidth}L{0.29\linewidth}Y}
\toprule
\tblgrouprow
\multicolumn{3}{c}{\tblhead{Public sources and terms}} \\
\midrule
\tblheaderrow
\tblhead{Dataset} & \tblhead{Public source} & \tblhead{License/terms} \\
\midrule
{\sffamily\bfseries HRBench-4K} & \texttt{DreamMr/HR-Bench} & Dataset-card field: ``other''; no standard open license declared \\
{\sffamily\bfseries V$^{*}$Bench} & \texttt{craigwu/vstar\_bench} & No dataset license declared on the official card; upstream terms apply \\
{\sffamily\bfseries WeMath} & \texttt{We-Math/We-Math} & CC BY-NC 4.0 \\
{\sffamily\bfseries MathVision-mini} & \texttt{MathLLMs/MathVision} & MIT (dataset card) \\
{\sffamily\bfseries MMStar} & \texttt{Lin-Chen/MMStar} & No standard dataset license declared; upstream sample terms apply \\
{\sffamily\bfseries RealWorldQA} & \texttt{xai-org/RealworldQA} & CC BY-ND 4.0 \\
{\sffamily\bfseries MMMU} & \texttt{MMMU/MMMU} & Apache-2.0 (dataset card); source-material disclaimer applies \\
\bottomrule
\end{tabularx}
\end{normaltable}
\end{table}

\begin{table}[H]
\centering
\caption{\textbf{Overview of experimental settings.} Common inference parameters are listed in Table~\ref{tab:core_settings}.}
\label{tab:study_provenance}
\begin{prosetable}
\begin{tabularx}{\linewidth}{L{0.17\linewidth}L{0.23\linewidth}L{0.25\linewidth}Y}
\toprule
\tblheaderrow
\tblhead{Study} & \tblhead{Models} & \tblhead{Evaluation scope} & \tblhead{Design} \\
\midrule
Main evaluation & Four base VLMs & Seven benchmarks & Vanilla, Append-only, and VLM-in-Sandbox \\
Matched controls & GPT-4.1-mini & Three benchmarks; 1,260 examples & Visibility $\times$ retention \\
Repeated evaluation & GPT-4.1-mini; Qwen3.5-9B & V$^{*}$Bench; RealWorldQA & Three evaluations per setting \\
Local efficiency & Qwen3.5-9B & Three benchmarks; 1,260 examples & Five settings with prefix caching on $8\times$H20 \\
Context scaling & Recorded trajectories & 100 trajectories & Fixed-trace replay with 4--32 artifacts \\
\bottomrule
\end{tabularx}
\end{prosetable}
\end{table}

We retain sample-level predictions, tool trajectories, and visual-state transitions for paired analysis and qualitative inspection.

\paragraph{Software environment.}
The sandbox uses a Linux Docker image with Python 3.11 and standard libraries for numerical computation, symbolic mathematics, plotting, image processing, and OCR. Local models are served separately through vLLM. The efficiency study uses $8\times$H20 with tensor parallelism 8 and request concurrency 1.

\subsubsection{Executor overhead and reliability}
\label{app:docker}

We measure the executor with local Qwen3.5-9B under Original Append-only on the complete three-benchmark collection. Each container is limited to 2 vCPUs and 4 GiB RAM, with networking disabled. A prebuilt image is reused; image preparation is excluded from per-example latency. Inputs are read-only, and each episode has an isolated writable workspace.

\begin{table}[H]
\centering
\caption{\textbf{Executor overhead and reliability} ($N=1{,}260$). Lifecycle time includes container creation, startup, and removal.}
\label{tab:docker_audit}
\begin{normaltable}
\begin{tabularx}{0.9\linewidth}{Y r}
\toprule
\tblheaderrow
\tblhead{Measurement} & \tblhead{Value} \\
\midrule
Container lifecycle / example & 0.82 s \\
Original Append-only E2E / example & 40.1 s \\
Lifecycle share of E2E & 2.0\% \\
Container lifecycle failures & 0 / 1,260 \\
Out-of-memory or resource-limit failures & 0 / 1,260 \\
Episodes with a failed tool call & 10 / 1,260 (0.8\%) \\
\bottomrule
\end{tabularx}
\end{normaltable}
\end{table}

These measurements characterize executor overhead and reliability, not a host-versus-container speed difference.

\section{Extended Quantitative Results}
\label{app:extended-results}

\subsection{Matched Controls and Component Sensitivity}

\begin{figure}[H]
\centering
\includegraphics[width=\linewidth]{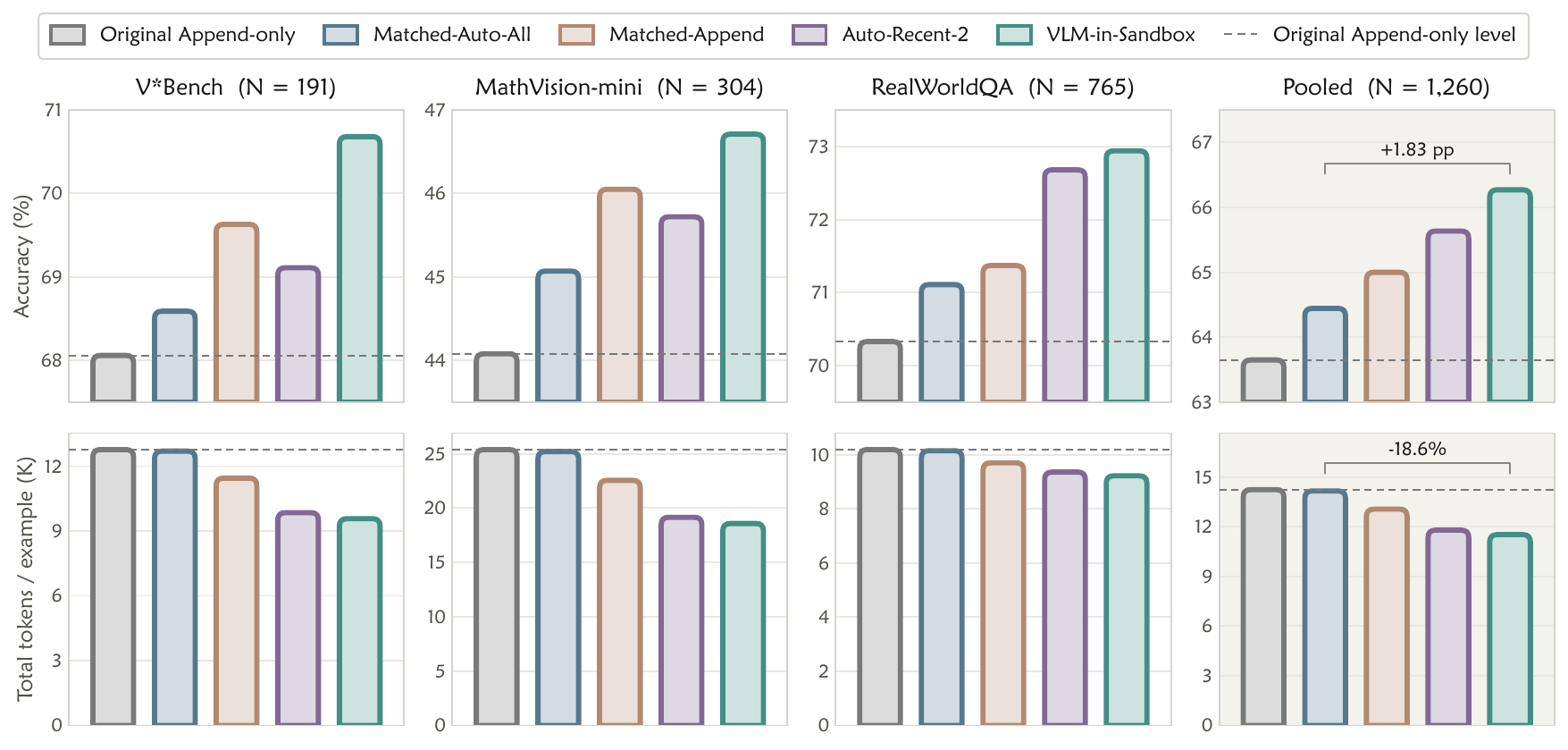}
\caption{\textbf{Full matched-control results.} GPT-4.1-mini on the three complete splits and their sample-size-weighted pooling, with accuracy (top row) and mean total tokens per example (bottom row) on separate axes. The four matched settings cross visibility (automatic or model-directed) with retention (all images or two active slots). Original Append-only uses the original append-only history instead of the shared prompt compiler, so it is a reference rather than a factorial cell; the dashed line marks its level in every panel. Brackets in the pooled column compare the compiler-matched control with VLM-in-Sandbox.}
\label{fig:matched_full}
\end{figure}

Figure~\ref{fig:matched_full} reports the complete three-dataset matched study. Accuracy and token cost respond to different factors: how much of the accuracy gain, and how much of the token saving, comes from visibility and how much from retention. Original Append-only is shown as the submitted reference but is not a cell of the $2\times2$ factorial because it uses the original append-only history. The other four configurations share the compiler and tool harness and vary visibility and retention, and the Original Append-only and VLM-in-Sandbox values reuse the corresponding main-evaluation results. Figure~\ref{fig:matched_controls} summarizes the factorial in the main text; the panels here add the reference setting and the pooled collection.

Reading the two rows separately decomposes the headline gain. Aligning the harness alone, from Original Append-only to Matched-Auto-All, accounts for $+0.79$ pp and a 0.6\% token change; the remaining $+1.83$ pp and essentially all of the 18.6\% token reduction come from managing visual state. The saving is uneven across splits, from 26.3\% on MathVision-mini, the most expensive, to 9.2\% on RealWorldQA, the cheapest.

The two factors then divide the work unevenly. Applied separately, selection and bounded retention give $+0.56$ and $+1.19$ pp, close to their joint $+1.83$ pp, whereas their separate token reductions of 7.8\% and 16.7\% fall well short of adding to the joint 18.6\%. Once the active context is capped, selection therefore changes mainly \emph{which} evidence is visible rather than how much of it, and it pays off in accuracy rather than in cost. Table~\ref{tab:matched_pairwise} tests these contrasts example by example rather than through aggregate rates: the two comparisons separated from zero at the 5\% level both add bounded retention, whereas adding selection at fixed capacity is not conclusive ($p=0.185$).

\begin{table}[H]
\centering
\caption{\textbf{Paired comparisons among matched settings.} Confidence intervals use 50,000 fixed-seed paired percentile-bootstrap resamples; $p$-values use exact McNemar tests. Token change is the right-hand setting relative to the left-hand setting.}
\label{tab:matched_pairwise}
\begin{normaltable}
\arrayrulecolor{black}
\renewcommand{\arraystretch}{1.08}
\begin{tabularx}{\linewidth}{@{}Y r r r r@{}}
\toprule
\tblhead{Comparison} & \tblhead{$\Delta$Acc.} & \tblhead{95\% CI} & \tblhead{$p$} & \tblhead{$\Delta$Tokens} \\
\midrule
Matched-Auto-All $\rightarrow$ VLM-in-Sandbox & \tblbest{+1.83 pp} & \tblbest{[0.40, 3.25]} & \tblbest{0.014} & \tblbest{$-18.6\%$} \\
Matched-Auto-All $\rightarrow$ Auto-Recent-2 & +1.19 pp & [0.00, 2.38]$^{\dagger}$ & 0.072 & $-16.7\%$ \\
Matched-Append $\rightarrow$ VLM-in-Sandbox & +1.27 pp & [0.24, 2.38] & 0.026 & $-11.8\%$ \\
Auto-Recent-2 $\rightarrow$ VLM-in-Sandbox & +0.63 pp & [$-0.16$, 1.51] & 0.185 & $-2.3\%$ \\
\bottomrule
\end{tabularx}
\tblnote{$^{\dagger}$The endpoint is displayed to two decimals. The bootstrap interval and exact McNemar test are distinct paired procedures and need not cross their decision thresholds identically.}
\end{normaltable}
\end{table}

The history-policy comparison in Section~\ref{subsec:context-growth} changes both the raw replay window and older-turn recap, rather than isolating recap alone. Figure~\ref{fig:slot_capacity} reports the accuracy and token trade-off for one, two, and four active slots.

\begin{table}[H]
\centering
\caption{\textbf{Agent-step budget sensitivity} on the matched-study collection ($N=1{,}260$), GPT-4.1-mini. Only the maximum number of agent steps varies; the default budget is highlighted. \emph{Exhausted} counts episodes that reach the limit before an accepted \texttt{submit} call.}
\label{tab:step_deadline}
\begin{normaltable}
\arrayrulecolor{black}
\renewcommand{\arraystretch}{1.08}
\begin{tabular}{@{}l r r r r@{}}
\toprule
\tblhead{Step budget} & \tblhead{Acc. (\%)} & \tblhead{Total tok.} & \tblhead{Mean steps} & \tblhead{Exhausted} \\
\midrule
10 & 65.71 & 10.74K & 4.3 & 26 (2.06\%) \\
\rowcolor{tabtop1purple}
20 & \tblbest{66.27} & 11.53K & 5.0 & 0 \\
30 & \tblbest{66.27} & 11.61K & 5.1 & 0 \\
\bottomrule
\end{tabular}
\end{normaltable}
\end{table}

The return on additional steps falls off sharply. Reducing the default budget to 10 steps saves 6.9\% of total tokens (11.53K to 10.74K per example) but costs 0.56 pp of pooled accuracy (66.27\% to 65.71\%), whereas raising it to 30 steps adds 0.7\% tokens and leaves the pooled estimate unchanged at 66.27\%. Mean trajectory length rises only from 5.0 to 5.1 steps over that increase, so the additional headroom is almost never used.

Termination is governed by the policy rather than by the cap. At 20 steps no episode reaches the limit, so every trajectory ends in an accepted \texttt{submit}; at 10 steps, 26 episodes (2.06\%) exhaust the budget, which places the accuracy difference in a small tail of long trajectories rather than in typical behavior. The two runs are not a pure truncation of one another, because the deadline reminder is defined relative to the remaining budget and therefore fires at a different absolute step; the comparison evaluates the step budget as a policy rather than isolating truncation. Removing the reminder from the 20-step configuration leaves 30 episodes (2.38\%) without an accepted \texttt{submit} call, whereas issuing it with three remaining steps, the submitted default ($d=3$ in Table~\ref{tab:core_settings}), reduces that count to zero, and with five remaining steps to three. Episodes stopped at the limit are still scored from a non-empty answer file when one exists, and termination status is recorded separately, so these counts measure protocol compliance rather than discarded examples.

\subsection{Paired Reliability and Complete Transitions}

\begin{table}[H]
\centering
\caption{\textbf{Complete outcome transitions to VLM-in-Sandbox.} Each row covers all seven benchmarks ($N=6{,}350$). Stable correct/wrong means both settings agree on correctness; rescue and regression mean wrong-to-correct and correct-to-wrong, respectively. The reference setting differs by model.}
\label{tab:complete_transitions}
\begin{normaltable}
\arrayrulecolor{black}
\renewcommand{\arraystretch}{1.08}
\begin{tabularx}{\linewidth}{@{}l Y r r r r@{}}
\toprule
\tblhead{Model} & \tblhead{Reference} & \tblhead{\makecell{Stable\\correct}} & \tblhead{Rescue} & \tblhead{Regression} & \tblhead{\makecell{Stable\\wrong}} \\
\midrule
GPT-4.1-mini & Original Append-only & 3,982 & \tblbest{302} & 142 & 1,924 \\
Qwen3.5-9B & Vanilla VLM & 4,224 & \tblbest{734} & 411 & 981 \\
\bottomrule
\end{tabularx}
\end{normaltable}
\end{table}

The primary GPT comparison is Original Append-only to VLM-in-Sandbox. We use a paired percentile bootstrap with 50,000 resamples and a fixed analysis seed. Pooled resampling is benchmark-stratified; the four HRBench cyclic variants are grouped by base question and sampled as a cluster. HRBench and the pooled collection use cluster-aware paired randomization tests, while other individual benchmarks use exact McNemar tests.

\begin{table}[H]
\centering
\caption{\textbf{GPT-4.1-mini paired results by benchmark.} Rescue and regression are defined relative to Original Append-only.}
\label{tab:paired_by_benchmark}
\begin{normaltable}
\arrayrulecolor{black}
\renewcommand{\arraystretch}{1.08}
\begin{tabular}{@{}l r r r r r r@{}}
\toprule
\tblhead{Dataset} & \tblhead{$N$} & \tblhead{Rescue} & \tblhead{Regression} & \tblhead{$\Delta$Acc.} & \tblhead{95\% CI} & \tblhead{$p$} \\
\midrule
HRBench-4K & 800 & 30 & 17 & +1.63 pp & [$-0.75$, 4.13] & 0.242 \\
V$^{*}$Bench & 191 & 14 & 9 & +2.62 pp & [$-2.09$, 7.85] & 0.405 \\
WeMath & 1,740 & 48 & 25 & +1.32 pp & [0.34, 2.30] & 0.0095 \\
MathVision-mini & 304 & 17 & 9 & +2.63 pp & [$-0.66$, 5.92] & 0.169 \\
MMStar & 1,500 & 121 & 43 & +5.20 pp & [3.53, 6.87] & $<0.001$ \\
RealWorldQA & 765 & 36 & 16 & +2.61 pp & [0.78, 4.44] & 0.0078 \\
MMMU & 1,050 & 36 & 23 & +1.24 pp & [$-0.19$, 2.67] & 0.117 \\
\midrule
\tblbest{All samples} & 6,350 & \tblbest{302} & \tblbest{142} & \tblbest{+2.52 pp} & \tblbest{[1.83, 3.21]} & \tblbest{$<0.001$} \\
\bottomrule
\end{tabular}
\end{normaltable}
\end{table}

The pooled analysis is the primary statistical result; the per-benchmark results show how the gains are distributed. Figure~\ref{fig:paired_reliability}(c--f) reports all three runs on complete V$^{*}$Bench and RealWorldQA splits at temperature 1.0. Trial 1 is the submitted Table~\ref{tab:main_results_acc} result; Trials 2--3 are independent repeats. Points are individual runs and horizontal bars are means, not confidence intervals.

The Qwen comparison yields 1.79 rescues per regression and 323 additional correct answers. RealWorldQA is the negative aggregate cell, with 53 rescues and 77 regressions. These outcomes motivate selective sandbox invocation on questions that a model can answer directly; they do not negate recovery on other examples.

\subsection{Scaling and Systems Efficiency}

The open-loop diagnostic in Figure~\ref{fig:context_growth} freezes trajectories and makes no model calls. It measures how compiled requests scale with stored visual artifacts, not task accuracy or closed-loop behavior. At recorded generation points, we add pixel-identical copies with distinct replay identifiers until the ledger contains 4, 8, 16, or 32 artifacts. Actions, observations, selections, and termination remain fixed; only the requests are recompiled. The copies increase stored evidence without introducing new visual information.

\begin{table}[H]
\centering
\caption{\textbf{Complete prefix-cache and latency accounting.} Qwen3.5-9B is served with vLLM on $8\times$H20 (tensor parallelism 8, concurrency 1), $N=1{,}260$. Token and E2E values are per example; TTFT is per request.}
\label{tab:cache_full}
\begin{densetable}
\arrayrulecolor{black}
\renewcommand{\arraystretch}{1.08}
\begin{adjustbox}{max width=\linewidth}
\begin{tabular}{@{}l r r r r r r r r@{}}
\toprule
& \multicolumn{5}{c}{\tblhead{Token workload / example}} & \tblhead{Prefix cache} & \multicolumn{2}{c}{\tblhead{Latency}} \\
\cmidrule(lr){2-6}\cmidrule(lr){7-7}\cmidrule(lr){8-9}
\tblhead{Setting} & \tblhead{Prompt} & \tblhead{Cached} & \tblhead{Uncached} & \tblhead{Completion} & \tblhead{Total} & \tblhead{Hit} & \tblhead{TTFT (s)} & \tblhead{E2E (s)} \\
\midrule
Original Append-only & 42.40K & 29.30K & 13.10K & 3.67K & 46.07K & 69.1\% & 2.46 & 40.1 \\
Matched-Auto-All & 40.90K & 27.40K & 13.50K & 3.55K & 44.45K & 67.0\% & 2.38 & 39.2 \\
Matched-Append & 35.32K & 23.63K & 11.69K & 2.91K & 38.23K & 66.9\% & 2.00 & 36.0 \\
Auto-Recent-2 & 32.10K & 20.30K & 11.80K & 2.75K & 34.85K & 63.2\% & 1.96 & 34.8 \\
\rowcolor{tabtop1purple}
VLM-in-Sandbox & \tblbest{31.39K} & 20.10K & \tblbest{11.29K} & \tblbest{2.63K} & \tblbest{34.02K} & 64.0\% & \tblbest{1.91} & \tblbest{34.1} \\
\bottomrule
\end{tabular}
\end{adjustbox}
\end{densetable}
\end{table}

Each setting begins from an empty cache. Prompt counts include multimodal placeholders; cached-prefix tokens are a subset of prompt tokens; prefix-KV hit ratio is cached-prefix tokens divided by prompt-input tokens; total is prompt plus completion. TTFT and E2E include multimodal preprocessing and vision encoding, while E2E also includes sandbox lifecycle and excludes offline scoring. Relative comparisons are restricted to this backend.

In the separate GPT-4.1-mini matched study, oversized requests affect 4, 2, 1, 0, and 0 of the 1,260 episodes for Original Append-only, Matched-Auto-All, Matched-Append, Auto-Recent-2, and VLM-in-Sandbox, respectively. This count includes episodes that later recover. Table~\ref{tab:docker_audit} reports executor lifecycle and failure accounting.

\clearpage
\section{Comparison with Related Approaches}
\label{app:scope}

Table~\ref{tab:interface_comparison} compares how related approaches represent, retain, and revisit intermediate visual evidence.

\begin{table}[H]
\centering
\caption{\textbf{Interface-level comparison with nearby visual-reasoning families.} ``Method-specific'' indicates that the property depends on the particular canvas, memory, or tool protocol rather than a general file-backed artifact lifecycle.}
\label{tab:interface_comparison}
\begin{densetable}
\begin{tabularx}{\linewidth}{L{0.18\linewidth}L{0.15\linewidth}L{0.18\linewidth}L{0.16\linewidth}L{0.15\linewidth}Y}
\toprule
\tblheaderrow
\tblhead{Family} & \tblhead{Persistent artifacts} & \tblhead{Identity / provenance} & \tblhead{Active visual budget} & \tblhead{Reactivation} & \tblhead{Training} \\
\midrule
{\sffamily\bfseries Visual scratchpads}~\cite{visualsketchpad} & Episode canvas/state & Method-specific & Method-specific & Update or regenerate & Varies \\
{\sffamily\bfseries PyVision}~\cite{pyvision} & Session artifacts & Session/tool-specific & No general slot bound & Tool view or rerun & No \\
{\sffamily\bfseries Multimodal memory}~\cite{vismem} & Model or memory state & Not a general file ledger & Method-specific & Retrieval/update & Usually trained \\
\tbloursrow
{\sffamily\bfseries VLM-in-Sandbox} & Sandbox filesystem & Ledger ID + provenance & Two active slots & \texttt{promote} by ID & No \\
\bottomrule
\end{tabularx}
\end{densetable}
\end{table}

Visual scratchpads make intermediate visual reasoning explicit through a canvas; PyVision demonstrates the value of dynamic Python tooling and session artifacts; multimodal-memory methods retain information in learned or method-specific memory. VLM-in-Sandbox contributes a different systems abstraction: generated files receive a persistent lifecycle of registration, provenance, bounded model-visible state, payload eviction, and model-directed reactivation. These approaches are complementary rather than mutually exclusive.

\section{A Failure of Intermediate Evidence Interpretation}
\label{app:evidence_failure}

\begin{figure}[H]
\centering
\includegraphics[width=\linewidth]{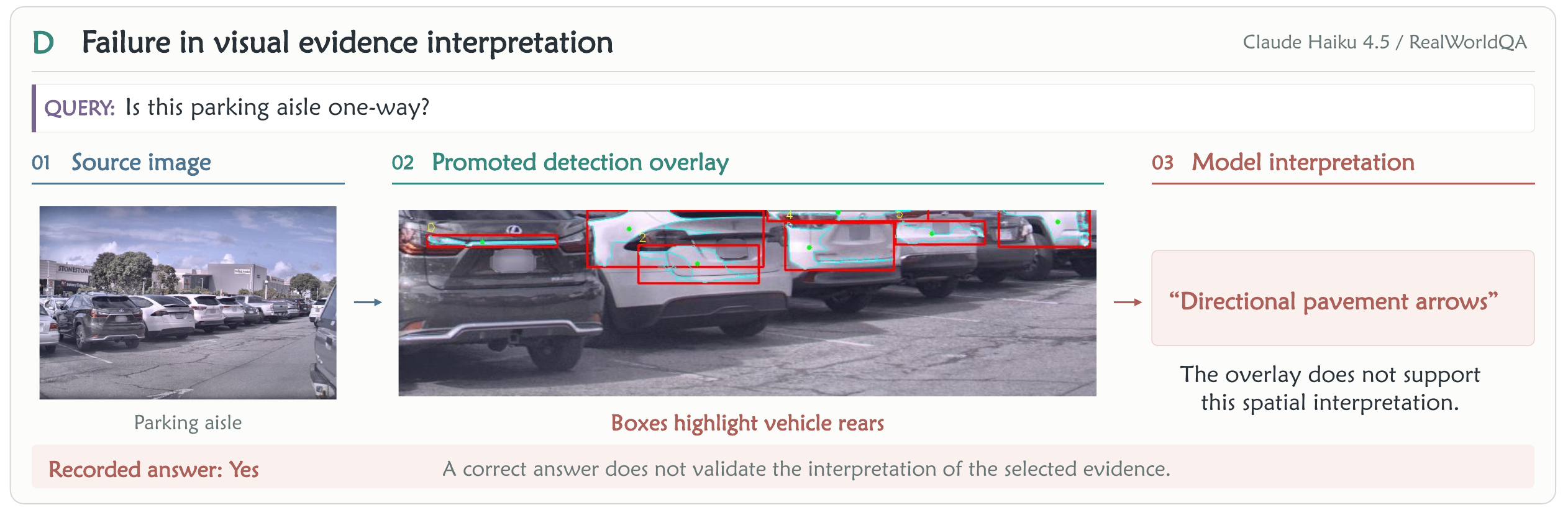}
\caption{\textbf{Correct answer with an unsupported visual rationale} (Claude Haiku 4.5, RealWorldQA). The promoted overlay highlights vehicles, whereas the model describes pavement arrows. The original detection overlay is retained to expose the mismatch between evidence and interpretation.}
\label{fig:case_evidence_failure}
\end{figure}

The parking-aisle example in Figure~\ref{fig:case_evidence_failure} illustrates a limitation distinct from answer accuracy. The model promotes a detection overlay and interprets its elongated contours as directional pavement arrows, but the boxes visibly lie on vehicle rears. The recorded final answer, yes, is correct for this RealWorldQA example; the stated visual rationale is nevertheless unsupported by the selected artifact. This example motivates checking the spatial grounding of intermediate evidence as well as its visibility. It is not counted as an answer-level regression.

\end{document}